\documentclass[10pt,twocolumn,letterpaper]{article}

\usepackage{iccv}

\usepackage{graphicx}
\usepackage{amsmath}
\usepackage{amssymb}
\usepackage{booktabs}
\usepackage{wrapfig}
\usepackage{times}
\usepackage{epsfig}
\usepackage{multirow}
\usepackage[ruled,vlined]{algorithm2e}
\usepackage{algpseudocode}
\usepackage{dsfont, bbm}

\usepackage{amsmath,amsfonts,bm}

\def\eqref#1{equation~\ref{#1}}

\def\1{\bm{1}}

\DeclareMathAlphabet{\mathsfit}{\encodingdefault}{\sfdefault}{m}{sl}
\SetMathAlphabet{\mathsfit}{bold}{\encodingdefault}{\sfdefault}{bx}{n}

\DeclareMathOperator*{\argmax}{arg\,max}
\DeclareMathOperator*{\argmin}{arg\,min}

\usepackage{xcolor}

\usepackage{enumitem}

\newcommand{\Eq}[1]{Eq.~(\ref{eq:#1})}
\newcommand{\eq}[1]{\Eq{#1}}

\newcommand{\fig}[1]{Fig.~\ref{fig:#1}}
\newcommand{\tab}[1]{Table~\ref{tab:#1}}

\newcommand{\alg}[1]{Algorithm~\ref{alg:#1}}

\usepackage{amsmath}

\newcommand{\RVC}{}   

\def\argmax{\operatornamewithlimits{\rm arg\,max}}

\def\argmin{\operatornamewithlimits{\rm arg\,min}}

\usepackage{graphicx}
\usepackage{amsmath}
\usepackage{amssymb}
\usepackage{booktabs}

\usepackage{xspace}

\usepackage[breaklinks=true,colorlinks,allcolors=black,bookmarks=false]{hyperref}

\iccvfinalcopy 

\def\iccvPaperID{12404} 

\ificcvfinal\fi

\begin{document}

%
\title{Differentiable Transportation Pruning}
\author{
Yunqiang Li$^{1}$ \ \ \  \ \ \   \ \   Jan C. van Gemert$^{2}$ \  \ \ \  \ \ \ \ Torsten Hoefler$^{3}$
\\ 
 \ \ \  \ \ \ Bert Moons$^{1}$  \  \  \ \ \ \  Evangelos Eleftheriou$^{1}$ \ \    Bram-Ernst Verhoef$^{1}$ 
\vspace{0.08in} \\
$^{1}$Axelera AI \ \ \ \ \ $^{2}$TU Delft \ \ \  \ \ $^{3}$ETH Zurich
\vspace{-0.04in}
}

\maketitle
\ificcvfinal\thispagestyle{empty}\fi

\begin{abstract}
Deep learning algorithms are increasingly employed at the edge. However, edge devices are resource constrained and thus require efficient deployment of deep neural networks. Pruning methods are a key tool for edge deployment as they can improve storage, compute, memory bandwidth, and energy usage. In this paper we propose a novel accurate pruning technique that allows precise control over the output network size. Our method uses an efficient optimal transportation scheme which we make end-to-end differentiable and which automatically tunes the exploration-exploitation behavior of the algorithm to find accurate sparse sub-networks. 
We show that our method 
achieves state-of-the-art performance compared to previous pruning methods
on 3 different datasets, using 5 different models,  across a wide range of pruning ratios, and with two types of sparsity budgets and pruning granularities. 
\end{abstract}
\vspace{-0.15in}
\section{Introduction}
\label{intr}
As the world is getting smaller, its edge is getting larger: more compute-limited edge devices are  used. To unlock deep learning on the edge, deep networks need to be efficient and compact. There is a demand for accurate networks that fit exactly in pre-defined memory constraints.  

Pruning is an effective technique~\cite{he2019filter, luo2020neural, ye2018rethinking} to find a sparse, compact model from a dense, over-parameterized deep network by eliminating redundant elements such as filters or weights while retaining accuracy. Pruning is a hard problem because  finding a good-performing sub-network demands selecting which part of a network to keep ---and which part to prune---, 
\begin{figure}
    \centering
\includegraphics[width=0.68\linewidth]{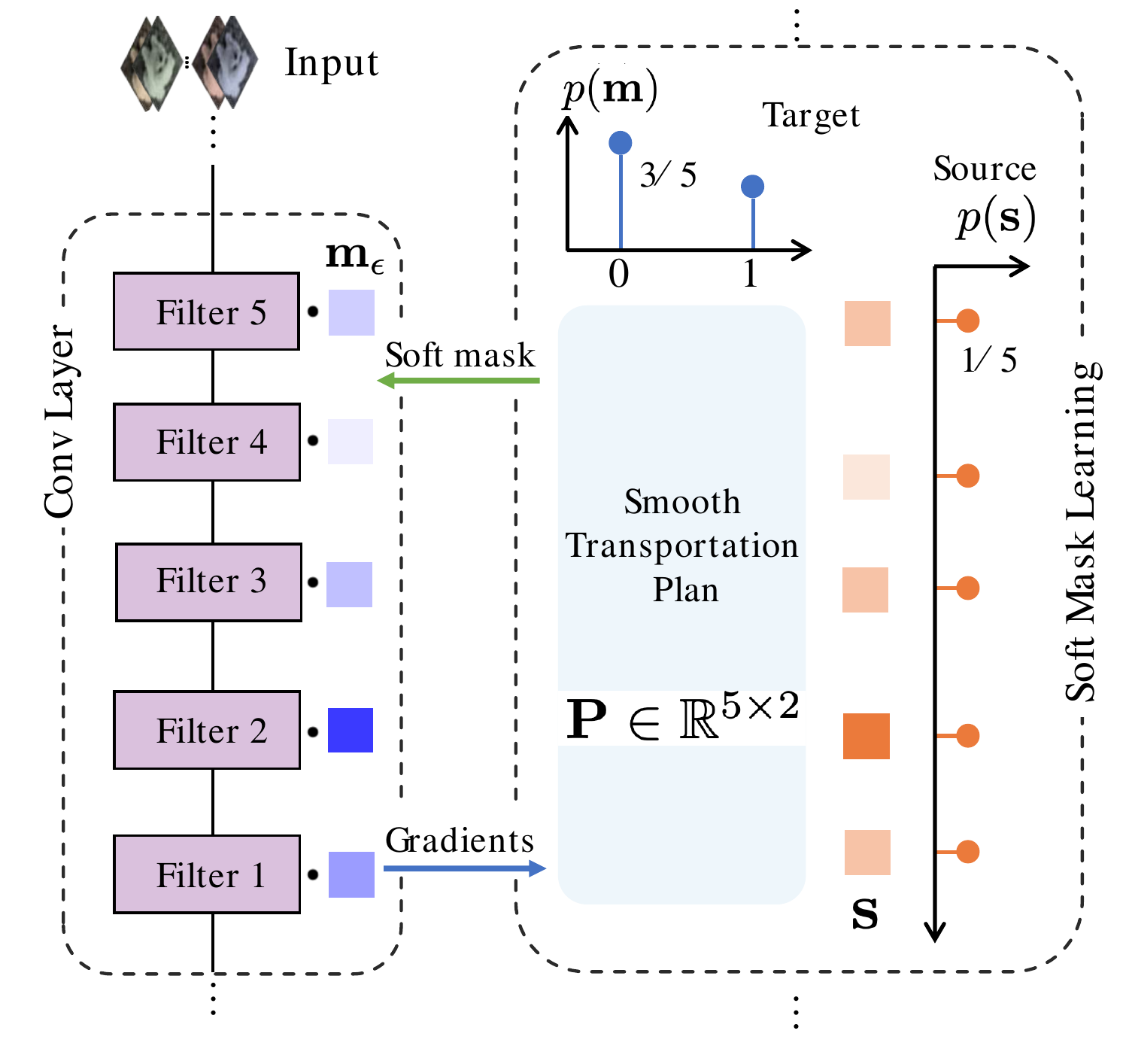}
    \vspace{-0.05in}
    \caption{ During training we multiply the filters in a layer with soft masks 
    $\mathbf{m}_{\epsilon}$. The soft masks are obtained by solving an optimal transport problem: minimizing a transportation cost between a uniform source distribution over trainable importance scores $\mathbf{s}$ 
    and a Bernoulli target distribution defined by sparsity ratio \ie ${3}/{5}$. 
    During training the soft masks gradually converge to hard masks.
    } 
    \label{fig:Methodoverview}
    \vspace{-0.2in}
\end{figure} 
\RVC{requiring an expensive search over  
a large discrete set of candidate architectures. 

Recent work~\cite{huang2018data, savarese2020winning, luo2020autopruner, kang2020operation} finds that optimizing  a continuously differentiable mask to sparsify network connection while simultaneously learning network parameters facilitates search space exploration
and enables gradient-based optimization. 
Prior methods~\cite{huang2018data, luo2020autopruner}, 
 however, control the sparsity, \ie pruned network size,
 by an additional penalty term in the loss function that is difficult to optimize and tune, and hampering a precise control over the sparsity. 
 

In this paper, as illustrated in~\fig{Methodoverview}, 
we propose a fully differentiable transportation pruning method that allows precise control of the output network size.
We draw inspiration from Xie et al.
(2020)~\cite{xie2020differentiable}
who formulate a soft top-$k$ operator based on entropic optimal transport.
Recent work~\cite{tai2022spartan}
shows good pruning results with the soft top-k operator. This soft top-k operator, however, is computationally expensive~\cite{ditschuneit2022auto} and relies on implicit, and therefore less accurate, gradients~\cite{eisenberger2022unified} for back-propagation. Instead, we speed up the soft top-k operator with an approximate bi-level optimization which is optimized using accurate automatic differentiation~\cite{paszke2017automatic}. Moreover, in contrast to ~\cite{xie2020differentiable, tai2022spartan} we can automatically tune the temperature hyper-parameter.

We make the following contributions:
{i)} A fully differentiable pruning method that allows precise control over output network size;
{ii)}  An efficient entropic optimal
transportation 
algorithm that significantly reduces the computational complexity
of bi-level optimization; 
{iii)}  We increase the ease-of-use of the algorithm by means of an automatic temperature annealing mechanism instead of a manually-chosen decay schedule;
{iv)}  We show state-of-the-art results  on 3 datasets with 5 models for structured and unstructured pruning using 
layer-wise and global sparsity budgets.







}

\section{Related Work}
\label{relatedwork}





\noindent \textbf{Pruning in deep learning.}
Network pruning makes models smaller by removing  elements that do not contribute significantly to accuracy. Pruning methods can be subdivided into methods that promote structured (\ie blocked)\RVC{~\cite{meng2020pruning, krashinsky2020nvidia}} or unstructured (\ie fine-grained) sparsity\RVC{~\cite{chijiwa2021pruning, frankle2018lottery, miao2021learning, vischer2021lottery, zhang2021lottery}}, see \cite{hoefler2021sparsity} for a review. Filter pruning is a form of structured sparsity by removing entire filters from the network's layers ~\cite{lin2020hrank, li2017pruning}. Filter pruning often achieves practical network compression and significant acceleration as entire feature maps are no longer computed. For these reasons, we here focus on filter pruning.


\noindent \textbf{Selection criteria for pruning.} 
Network pruning can be phrased as a form of  neural architecture search~\cite{liu2018rethinking}.
Such an architecture search may involve training and evaluating several random subnetworks based on leave-some-out approaches~\cite{changpinyo2017power, suzuki2001simple, li2022revisiting}, but this can be quite expensive for large models. 
More efficient approaches assign an importance score as selection criteria.
This score can be based on  L1/L2 norm of weights~\cite{ye2018rethinking},  geometric median~\cite{he2019filter} of filters,   Kullback–Leibler divergence~\cite{luo2020neural},
\RVC{Wasserstein barycenter of channel probability vectors~\cite{shen2020cpot},}
 prediction error~\cite{molchanov2019importance}, or  empirical sensitivity of a feature map~\cite{liebenwein2019provable}. These methods first prune a pretrained model based on the importance scores and subsequently fine-tune the pruned model so the model can adapt to the reduced set of parameters. Once pruned, however, the model is not allowed to explore other subnetworks. Here, we also use importance scores, which we learn as latent variables from which soft masks are computed through optimal transport. The soft mask is automatically annealed to a hard mask during training, allowing the model to explore different subnetworks in the process.



\noindent \textbf{Differentiable  continuous relaxation.} 
Recent work aims to select the best sub-architecture by training a binary mask on the network weights~\cite{gao2020discrete}. This binary mask, however, is non-differentiable and applying a hard binary mask during the forward pass may impair network performance~\cite{huang2018data}.
Other methods approximate binary masks using a sigmoid function, which gradually converges to a binary step function by means of a decaying temperature hyper-parameter~\cite{huang2018data, kang2020operation, luo2020autopruner, savarese2020winning}. To control the sparsity ratio, these methods usually introduce a sparsity penalty term in the loss function, which requires an  additional difficult-to-tune hyper-parameter if an exact pruning ration is required. 
Our method mitigates these issues as it allows for an exact pruned network size using optimal transport and decays the temperature variable automatically.

\RVC{ \noindent \textbf{Learning soft mask via entropic optimal transport.} 
Our method draws inspiration from the work of Xie et al. (2020)~\cite{xie2020differentiable}, who formulate a soft top-$k$ operator based on entropic optimal transport~\cite{peyre2019computational}, optimized via Sinkhorn’s algorithm~\cite{cuturi2013sinkhorn}.  The authors apply their method to predictive modeling applications such as information retrieval. As remarked by~\cite{ditschuneit2022auto}, this method can be computationally expensive if applied to pruning, as hundreds of Sinkhorn iterations are needed in the forward pass of each update step. In this paper we phrase network pruning as an efficient optimal transport problem. Compared to ~\cite{xie2020differentiable} our method significantly reduces the number of Sinkhorn iterations required in the forward pass. Furthermore, in contrast to~\cite{xie2020differentiable}, who employ implicit gradients~\cite{eisenberger2022unified}, our method relies on accurate automatic differentiation~\cite{paszke2017automatic}. Finally, we replace the manually-chosen decay schedule of the entropic hyperparameter in~\cite{xie2020differentiable} with an automatic schedule, thereby simplifying the optimization process. 


}

\RVC{
\noindent \textbf{Bregman divergence based proximal method.} 
Previous work~\cite{kosowsky1994invisible, schmitzer2019stabilized} has applied to optimal transport the 
proximal point algorithm~\cite{parikh2014proximal} based on the Bregman divergence~\cite{censor1992proximal} to alleviate the potential numerical instability of Sinkhorn iterations. 
Xie \etal~\cite{xie2020fast} use the proximal algorithm to optimize optimal transport in deep generative models~\cite{arjovsky2017wasserstein, genevay2018learning} and achieve good performance, but still 
performs a large amount of proximal point iterations per mini-batch step, making it
computational inefficient for our pruning purposes. In addition, its back-propagation pass~\cite{xie2020fast} relies on the envelope theorem~\cite{afriat1971theory} that does not go into proximal point iterations to accelerate training, which causes inaccurate gradients.
Instead, we only use a single proximal point iteration, \ie a single Sinkhorn iteration, per mini-batch step and compute gradients via automatic differentiation~\cite{paszke2017automatic}.



}

\RVC{\noindent\textbf{Budget aware pruning.} 
Budgeted pruning  focuses on compressing the network subject to the prescribed explicit resource constraints, 
such as targeted amount
of layer-wise filters~\cite{lemaire2019structured}, global filters~\cite{tiwari2021chipnet}, FLOPs~\cite{li2020eagleeye, guo2020dmcp}, or execution latency~\cite{shen2022structural, Humble2022pruning}.
Our method can directly control the pruned network size given a sparsity budget. Moreover, we show how it can be easily extended to general budgeted pruning by learning budget-constrained sparsity across
layers~\cite{ning2020dsa}.
Our technique improves the accuracy while using less hyperparameters, \eg ChipNet~\cite{tiwari2021chipnet} adds
6 extra hyperparameters while we only introduce a single temperature hyperparameter, which eases hyperparameter tuning effort.

 }


\section{ 
Differentiable Transportation Pruning}
\label{methods}

\begin{figure*}[t]
\centering
\begin{tabular}{cc}
\includegraphics[width=0.99\textwidth]{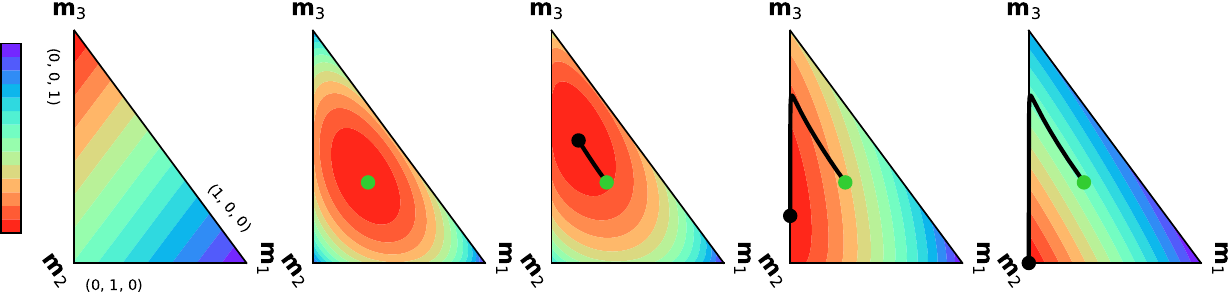}  
 \\
(a) \RVC{no entropic relaxation}
 \ \  \ \  
   (b)  $\ell = 10^0$  \  \ \ \ \ \ \ \ \ \ \ \  \ \ \ \ \ 
    (c) $\ell = 10^1$   \ \ \ \ \ \ \  \ \ \ \ \  \ \    \ \ \ \  \   (d) $\ell = 10^2$ \ \ \ \ \  \ \ \ \  \  \ \ \ \ \  \ \ \     (e) $\ell = 10^3$ 
      \vspace{0.03in}
   \\
         \includegraphics[width=1\textwidth]{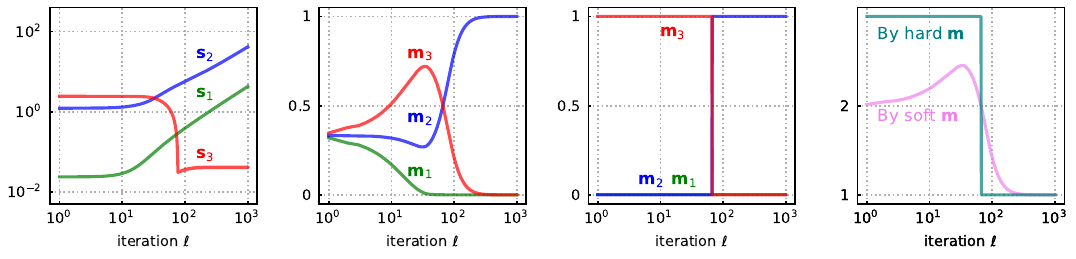}
& \\
 \ \ \ \
(f) importance score \  \ \ \ \ \ \   \ \ \ \  \ \  \ \ \ \ \   \ \
   (g) soft mask \ \  \ \ \ \ \ \ \  \ \ \ \ \  \ \  \ \ \ \ \ 
 \ \  \ \ \ \ \  (h) hard mask \ \  \ \ \ \ \ \ \  \ \ \ \ \  \ \  \ \ \ \ \   \ \   (i) training loss 
    \end{tabular}
    \caption{ 
\textbf{Illustrative example:} The figure is best understood after \eq{optsolutionSoft} and it illustrates how our method can prune 2 weights out of 3 by learning a binary mask $ \mathbf{m} \in \{0, 1\}^3$. We can visualize in 2D because the simplex $\small \sum_3$ can be shown as a triangle in two dimensions. In this example we chose to favor the 2nd weight, and thus define some loss on the weights as $\mathcal{L}_\text{train} = \mathbf{w}^{\mathsf{T}}\mathbf{m} $  where we define $\mathbf{w} = [2, 1, 3]$.
To keep just a single weight, we set $k$ to 1, yielding the constraint ${\sum_{i=1}^3 \mathbf{m}_i}  = 1$ from  \eq{formulate_pruningnew}, where the optimal solution is $\mathbf{m} = (0, 1, 0)$. \textit{The top row (a)-(e)} illustrates the solution space, where the corners of the triangle are the hard masks: $(1, 0, 0)$, $(0, 1, 0)$ and $(0, 0, 1)$. 
\RVC{Specifically, (a), (b) denote the sparse and non-sparse solution space without or with  entropy regularization at $\ell = 1$.} 
The color bar indicates low (red) and high (violet) loss values. A green dot in the center is the initial position; and the black dot is the current position. 
\textit{the bottom row (f)-(i)}:  training curve of importance score, soft mask, discrete mask and training loss for one thousand training steps. The hard mask in (h) is obtained by \eq{optsolution}. \textit{(a)}: 
the discrete optimal transportation loss at $\ell = 1$ just after initializing the importance scores $\mathbf{s}$. Note the ordering of $\bf s_1, s_2, s_3$ in (f) at $\ell = 1$, which explains the low loss value for $\bf m_3$.
\textit{(b)-(e)}: Snapshots at iterations $\ell = 10^0, 10^1, 10^2, 10^3$ after making the discrete optimal transport problem differentiable by entropy regularization in \eq{entropyopt2} where $\varepsilon = 10$. Note how the black dot moves to the correct solution.
Note the training loss $\mathcal{L}_\text{train}$ in (i), as it pushes $\mathbf{s}_2$ towards top-1 in (f), and minimizing transportation loss pushes the soft mask $\mathbf{m}_2$ towards 1 in (g). The minimum value of the training loss is $1$, due to how we chose the dot product loss $\mathcal{L}_\text{train}$.
}
\label{fig:toy_example}
\end{figure*}

Given a neural network $f(\mathbf{x}; \mathbf{w})$ with input  $\mathbf{x}$ and a set of $n$ filters forming the columns of weight matrix $\mathbf{w} \in \mathbb{R}^{m \times n}$, filter pruning removes filters by applying a binary mask $\mathbf{m} \in \{0, 1\}^n $ to the $n$ filters ($1=\text{kept}$, $0=\text{pruned}$). 

If we need to keep exactly $k$ out of $n$ filters, then the sparse learning problem can be formulated as: 
 \begin{equation}
\begin{split}
& \mathop{\argmin}_{\mathbf{m}, \mathbf{w}}\ \ 	\mathcal{L}_\text{train} \big(f(\mathbf{x};  \mathbf{w}, \mathbf{m})\big),
 \\
 & \ \ \ \text{s.t.} \ \ \   
 \sum \nolimits_{i=1}^n \mathbf{m}_i = k,
\  \   \mathbf{m} \in \{0, 1\}^n,
\end{split}
\label{eq:formulate_pruningnew}
\end{equation}
where $\mathcal{L}_\text{train}$ is  the training loss, \eg \ cross entropy loss.

The binary mask $\mathbf{m}$ is discrete and therefore difficult to optimize with gradient descent. To overcome this, we make the binary mask  depend on a continuous latent importance score $\mathbf{s}$ which can be optimized by gradient descend. By sorting these importance scores we allow exact control over the pruned output network size by keeping only the top-$k$ importance scores. 
Specifically, the top-$k$ operator 
 first sorts the importance scores $\mathbf{s}$, and then assigns the top $k$ importance scores to $1$ and the remaining importance scores to $0$, that is:
  \begin{equation}
\mathbf{m}_i  =
\begin{cases}
1,  & \mathrm{if \  } \mathbf{s}_i  \ \mathrm{ is \  in \ the \ top}$-$k \ \mathrm{of } \ \mathbf{s},
\\
0, & \mathrm{otherwise}.
\end{cases}
\label{eq:formulate_topk}
\end{equation}
The top-$k$ operator can be parameterized in terms of the solution of an optimal transport problem~\cite{ xie2020differentiable}. This allows us to parameterize the pruning mask in terms of optimal transport, which is the problem of efficiently moving probability mass from a source distribution to a target distribution.
In our case, we know the target Bernoulli distribution over the binary values, which is given by the fraction of the $k$ filters out of the $n$ total filters to keep: \RVC{$P(1)=\frac{k}{n}$ and $P(0)=\frac{n-k}{n}$}. Then, the optimization problem reduces to minimizing the transportation costs of moving probability mass from the trainable importance scores $\mathbf{s}$ to this target distribution. See \fig{Methodoverview} for an illustration.



The two probability distributions used in optimal transport are over the importance scores $\mathbf{s}$ and 
over the two binary values. 
We define them by considering two discrete measures  $\mathbf{a} = \sum_{i = 1}^n \mathbf{a}_i \delta_{\mathbf{s}_i}$ and 
$\mathbf{b} = \sum_{j = 1}^2 \mathbf{b}_j \delta_{\mathbf{q}_j}$ supported on $\{\mathbf{s}_i\}_{i = 1} ^n$ and $\{\mathbf{q}_j:\{0, 1\}\}$ respectively   where $\delta_x$ is the  Dirac  at location $x$, 
\RVC{intuitively a unit of mass which is infinitely
concentrated at location $x$,}
the $\mathbf{a}_i$ and $\mathbf{b}_j$ are the probability mass. 
Following~\cite{xie2020differentiable}, we define $\mathbf{a}$ as \RVC{an empirical discrete} uniform distribution (source)  with $\mathbf{a}_i = {1} / n$. 
Our binary mask variable is distributed as
$\mathbf{b} = [1-(k/n), k/n]^{\mathsf{T}}$~(target). 

These two distributions now allow  optimal transportation optimization where probability mass is moved according to the cost it takes to move one unit of probability mass.
We define the cost matrix $\mathbf{C} \in \mathbb{R}^{n \times 2}$ which aligns $n$ filters (source) to the 2 binary options (target). The elements of the cost matrix are squared Euclidean distances and are denoted as:
  \begin{equation}
\mathbf{C}_{i1} = \mathbf{s}_i^2, \ \mathbf{C}_{i2} = (\mathbf{s}_i - 1)^2, \ i = 1, 2,...,n
\label{eq:cost_C}
\end{equation}
where the targets are binary (0 or 1) and thus the target value  for $\mathbf{C}_{i1}$ is $0$, and the target value for $\mathbf{C}_{i2}$ is~$1$. 

With the source distributions, target distribution  and cost matrix defined,
the optimal transportation plan can be formulated as: 
  \begin{equation}
\mathbf{P}^*= \mathop{\argmin}_{\mathbf{P} \in\mathcal{U}(\mathbf{a}, \mathbf{b})}	\left \langle \mathbf{C}, \mathbf{P}\right \rangle, 
\label{eq:opt}
\end{equation}
where $\mathcal{U}(\mathbf{a}, \mathbf{b}) = \{\mathbf{P}\in \mathbb{R}^{n \times 2}: \mathbf{P} 
 \mathds{1}_2 = \mathbf{a}, \  \mathbf{P}^\mathsf{T} \mathbbm{1}_n = \mathbf{b}\}$ is a set of coupling measures satisfying marginal constraints $\mathbf{P} 
 \mathbbm{1}_2 = \mathbf{a}$ and $   \mathbf{P}^\mathsf{T} \mathbbm{1}_n = \mathbf{b} 
$,
$\mathds{1}_n$ and $\mathds{1}_2$  denote vectors with $n$ ones and 2 ones,
and   
$\mathbf{P}\in \mathbb{R}^{n \times 2}$ denotes the general probabilistic coupling matrix, \ie the transportation plan, 
\RVC{and $\left \langle \cdot, \cdot \right \rangle$ is the Frobenius dot-product.}

The optimal transportation plan  $\mathbf{P}^*$ can be computed  as:
 \begin{equation}
\mathbf{P}^*_{\sigma_i, 1}  =
\begin{cases}
1/n,  & \mathrm{if \  } i \leq k \\
0, & \mathrm{otherwise}
\end{cases}, 
\mathbf{P}^*_{\sigma_i, 2} = 
\begin{cases}
0, & \mathrm{if \  } i \leq k \\
1/n, & \mathrm{otherwise}
\end{cases}
\label{eq:hardassign}
\end{equation}
 with $\sigma$  being the sorting permutation, \ie \ $\mathbf{s}_{\sigma_1} < \mathbf{s}_{\sigma_2} <\cdot\cdot\cdot < \mathbf{s}_{\sigma_n}$. 
 Given $\mathbf{P}^*$, the binary mask can be parameterized as a function of the optimal transport plan: 
 \begin{equation}
\mathbf{m} = n\mathbf{P}^*\cdot[0, 1]^{\mathsf{T}}.
\label{eq:optsolution}
\end{equation}


\noindent \textbf{Differentiable transportation relaxation.}
So far, the derived hard mask in \eq{optsolution} 
\RVC{is not differentiable everywhere}
with respect to importance scores, which can be seen from the sorting permutation applied in \eq{hardassign} and also in the lack of a gradient when using masking with a hard zero. 

To make the method differentiable, we derive a soft mask by smoothing the optimal transportation plan with entropic regularization. By adding an entropy regularization term we extend the solution space by allowing non-sparse solutions of the transportation plan. That is, we now allow soft masks.

Define the discrete entropy of a coupling matrix as $\mathcal{H}(\mathbf{P}) = -\sum_{ij}\mathbf{P}_{ij}(\text{log}(\mathbf{P}_{ij}) - 1)$, an entropic regularization is used  to  obtain smooth solutions to the original transport problem in \eq{opt}:
   \begin{equation}
\mathbf{P}^{*}_{\varepsilon}= \mathop{\argmin}_{\mathbf{P} \in\mathcal{U}(\mathbf{a}, \mathbf{b})}	\left \langle \mathbf{C}, \mathbf{P}\right \rangle - \varepsilon \mathcal{H}(\mathbf{P}).
\label{eq:entropyopt2}
\end{equation}
This objective is an $\varepsilon$-strongly convex function
which has a unique solution. 
As $\varepsilon\to0$ the unique solution $\mathbf{P}_{\varepsilon}$ of (\ref{eq:entropyopt2}) converges to the original optimal sparse solution of (\ref{eq:opt}), \ie with hard masks. 


Our soft masks $\mathbf{m_{\epsilon}}$ are now obtained by inserting  $\mathbf{P}^*_{\varepsilon}$ into (\ref{eq:optsolution}), as:
\begin{equation}
\mathbf{m_{\epsilon}} = n\mathbf{P}^*_{\varepsilon}\cdot[0, 1]^{\mathsf{T}}.
\label{eq:optsolutionSoft}
\end{equation}

\noindent \textbf{Interlude: illustrative example.} We have now described the main ingredients of our method. This allows us to give an illustrative example of the main setting before we describe how to solve the optimization problem. In \fig{toy_example} we show a 2D visualization for pruning $n=3$ importance scores to just a single score.

\noindent \textbf{Dual formulation for optimization.} To facilitate the optimization problem in (\ref{eq:entropyopt2}), we make use of Lagrangean duality. That is, by introducing two dual variables $\mathbf{f}\in \mathbb{R}^{n}$ and $\mathbf{g}\in \mathbb{R}^{2}$ for each marginal constraint of the transport plans in  $\mathcal{U}(\mathbf{a}, \mathbf{b})$, 
the Lagrangian of (\ref{eq:entropyopt2}) is optimized, resulting in 
 an optimal $\mathbf{P}_{\varepsilon}$ to the regularized problem given by:
    \begin{equation}
\mathbf{P}_{\varepsilon} = e ^{{\mathbf{f}}/{\varepsilon}} \odot e ^{-{\mathbf{C}(\mathbf{s})}/{\varepsilon}} \odot e ^{{\mathbf{g}}/{\varepsilon}}.
\label{eq:lagrange}
\end{equation}
where  $\odot$ is element wise product (see details in \ref{Optimal_coupling}).
Note that the $\mathbf{P}_{\varepsilon}$ in \eq{lagrange} is differentiable with respect to $\mathbf{s}$.


\begin{algorithm*} 
\caption{\RVC{Differentiable Transportation Pruning}}\label{alg:algrithorm}
Probabilities $\mathbf{a}={\mathbbm{1}_n}/{n}$, $\mathbf{b} = [p, 1-p]^{\mathsf{T}}$ on supports $\{\mathbf{s}_i\}_{i = 1} ^n$, $\{\mathbf{m}_j\}_{j = 1} ^2$,
  weights: $\mathbf{w}$, training loss: $\mathcal{L}_\text{train}$, 
  training iterations: $L$,  learning rate:  $\alpha$, initialization: $\varepsilon$, 
 $\mathbf{g}^{(1)} \gets \mathbbm{1}_2 $, $\mathbf{P}^{(1)} \gets \frac{1}{n}\mathbbm{1}_n \mathbbm{1}_2^{\mathsf{T}}$
{
 
 \For  {$ \ell = 1, 2, ... , {L}$}{
 Cost matrix $\mathbf{C}$ with element $\mathbf{C}_{ij} = (\mathbf{s}_i^{(\ell)} - \mathbf{m}_j)^2$; \ \   Gibbs kernel $\mathbf{K} = e^{-\frac{\mathbf{C}}{\varepsilon}} \odot \mathbf{P}^{(\ell)}$

 Update dual variables:
 $\mathbf{f}^{(\ell+1)} = \varepsilon \ \text{log} \ \mathbf{a}  - \varepsilon \ \text{log} \ (\mathbf{K}e^{\mathbf{g}^{(\ell)}/\varepsilon} )$; \ \
 $\mathbf{g}^{(\ell+1)} = \varepsilon \ \text{log} \ \mathbf{b}  - \varepsilon \ \text{log} \ (\mathbf{K}^{\mathsf{T}}e^{\mathbf{f}^{(\ell + 1)}/\varepsilon} )$

Update  transportation plan $\mathbf{P}^{(\ell+1)} = e^{{\mathbf{f}^{(\ell+1)}}/{\varepsilon}}  \odot \mathbf{K}  \odot e^{{\mathbf{g}^{(\ell+1)}}/{\varepsilon}}$

   Update model variables with SGD:
\ \ $\mathbf{s}^{(\ell+1)} = \mathbf{s}^{(\ell)} - \alpha   
\nabla_\mathbf{s}\mathcal{L}_\text{train} $; \ \  $\mathbf{w}^{(\ell+1)} = \mathbf{w}^{(\ell)} - \alpha   \nabla_\mathbf{w}\mathcal{L}_\text{train}$
}

 Derive  pruned architecture based on soft mask $\mathbf{m} = n\mathbf{P}^{({L})}\cdot[0, 1]^{\mathsf{T}}$ 
 for finetuning.
 }
\end{algorithm*}

\noindent \textbf{Bi-level optimization.}
Now that we have relaxed the problem, we aim to jointly learn the dual
variables $\mathbf{f}$, $\mathbf{g}$ and the model variables $\mathbf{s}$, $\mathbf{w}$. 
The dual variables are optimized by minimizing the Sinkhorn divergence as in \eq{entropyopt2}.  
This is equivalent to maximizing its dual problem, denoted as   $\mathcal{L}_\text{dual}^{\varepsilon} (\mathbf{f}, \mathbf{g}, \mathbf{s}) = \left \langle \mathbf{f}, \mathbf{a}\right \rangle
+ \left \langle \mathbf{g}, \mathbf{b}\right \rangle
- \varepsilon \left \langle e ^{{\mathbf{f}}/{\varepsilon}},  e ^{-{\mathbf{C}{(\mathbf{s})}}/{\varepsilon}} \cdot e ^{{\mathbf{g}}/{\varepsilon}}\right \rangle$ (See details in \ref{Dual_problem}). Finally the model variables are obtained by minimizing the training loss $\mathcal{L}_\text{train}$.

This implies a
\textit{bi-level optimization problem}~\cite{colson2007overview}, 
\ie an optimization problem which contains another optimization problem as a constraint.
The model variables $\mathbf{s}$, $\mathbf{w}$ appear as the upper-level variables, and the dual variables $\mathbf{f}$,  $\mathbf{g}$ as the lower-level variables. We rewrite the original problem in \eq{formulate_pruningnew}  as the following bi-level optimization problem:
\begin{align}
& \mathop{\min}_{\mathbf{s}, \mathbf{w}}\ \ 	\mathcal{L}_\text{train} \Big(f \big(\mathbf{x}; \mathbf{w}, \mathbf{m} (\mathbf{P}^*_{\varepsilon}(\mathbf{s}))\big) \Big),
 \label{eq:upper} \\
& \ \ \  \text{s.t.} \ \ \  \mathbf{f}^*, \mathbf{g}^*  =  	\mathop{\argmax} \nolimits_{\mathbf{f},  \mathbf{g}}  \   \mathcal{L}_\text{dual}^{\varepsilon} (\mathbf{f}, \mathbf{g}, \mathbf{s}),\label{eq:lowerl} 
	\end{align}
where $\mathbf{P}^*_{\varepsilon}(\mathbf{s})$ is obtained by inserting the outcome from (\ref{eq:lowerl}) into (\ref{eq:lagrange}), and the mask $\mathbf{m}_{\varepsilon}$  in (\ref{eq:upper}), dubbed soft mask, is computed by inserting  $\mathbf{P}^*_{\varepsilon}(\mathbf{s})$ into~\eq{optsolutionSoft}.



Nonetheless, this bi-level optimization problem is expensive to train with Sinkhorn's algorithm:
 Firstly, per mini-batch the inner optimization  in \eq{lowerl} needs to perform Sinkhorn's iterative algorithm for hundreds of iterations to converge, which is computationally inefficient. 
Secondly, during the back-propagation pass, the gradient of $\mathbf{P}^*_{\varepsilon}(\mathbf{s})$
 with respect to the importance scores  $\mathbf{s}$ is computed by differentiating~\cite{paszke2017automatic} through all Sinkhorn iterations, which is expensive (see details in \ref{For_back_sinkhorn}). Finally, we have also  
introduced an  additional hyperparameter $\varepsilon$ to decay, which makes the problem even harder to optimize.

\noindent \textbf{Approximate bi-level optimization.}
\RVC{We note that in our bi-level optimization the target distribution is a fixed Bernoulli distribution with parameter defined by the sparsity. The source distribution changes only when the trainable importance score updates via SGD. 
This differs from previous approaches~\cite{xie2020differentiable, xie2020fast} involving a stochastic  sampling of the target or source distribution and where the inner optimization in \eq{lowerl} is expected to converge per mini-batch step. 
We therefore propose to
approximate $\mathbf{P}^*_{\varepsilon}(\mathbf{s})$ in \eq{lowerl} using  only a single 
Sinkhorn step, \ie, without 
optimizing the inner optimization problem until convergence per mini-batch step. 

Given the $\ell$-th mini-batch update step, we approximate the training loss in~\eq{upper} with:
 \begin{equation}
\begin{split}
&\mathcal{L}_\text{train} \Big(f \big(\mathbf{x}; \mathbf{w}, \mathbf{m} (\mathbf{P}^*_{\varepsilon}(\mathbf{s}))\big) \Big)
\\
 \approx 
 &\mathcal{L}_\text{train} \Big(f \big(\mathbf{x}; \mathbf{w}, \mathbf{m} (\mathbf{P}^{(\ell+1)}_\varepsilon(\mathbf{s}))\big) \Big),
\end{split}
\label{eq:approximate}
\end{equation}
where $\mathbf{P}^{(\ell+1)}_\varepsilon(\mathbf{s})$ is obtained by applying a single Sinkhorn update step. 
Similar techniques for solving bi-level optimization have been used in architecture search~\cite{liu2019darts}, hyperparameter tuning~\cite{luketina2016scalable} and meta-learning~\cite{finn2017model}.}

\RVC{
To enhance the convergence guarantee  using just a \textit{single Sinkhorn inner  iteration}, 
we use generalized proximal point iteration
 based on Bregman divergence~\cite{schmitzer2019stabilized, xie2020fast} (see details in \ref{Bregman}), 
 which can be viewed as applying iteratively the Sinkhorn algorithm~\cite{peyre2019computational} with a
$e^{-\frac{\mathbf{C}}{\varepsilon/\ell}}$ 
kernel (see details in \ref{proximal_sinkhorn}).
Thus, a proximal
point iteration keeps track of the previous Sinkhorn update like Momentum~\cite{sutskever2013importance} that incorporating a moving average of previous gradients, which accelerates the convergence and improves optimization stability.  
We therefore compute  $\mathbf{P}^{(\ell+1)}_\varepsilon(\mathbf{s})$ in \eq{approximate} as,
  \begin{align}
\mathbf{P}^{(\ell+1)}_\varepsilon(\mathbf{s}) &= e ^{\frac{\mathbf{f}^{(\ell+ 1)}}{\varepsilon}}  \odot \big(e^{-\frac{\mathbf{C}(\mathbf{s})}{\varepsilon}} \odot \mathbf{P}^{(\ell)}\big)  \odot e ^{\frac{\mathbf{g}^{(\ell + 1)}}{\varepsilon}},
\label{eq:decayauto}
\end{align}
using the transportation plan $\mathbf{P}^{(\ell)}$ from the previous step. 
Let $\mathbf{K} = e^{-{\mathbf{C}(\mathbf{s})}/{\varepsilon}} \odot \mathbf{P}^{(\ell)}$ denote the Gibbs kernel in the Sinkhorn algorithm, and the dual variables can be computed using a single Sinkhorn iteration, \ie a proximal point iteration with block coordinate ascent as (see details in \ref{sinkhorn}): 
\begin{align}
  &\mathbf{f}^{(\ell+1)} = \varepsilon \ \text{log} \ \mathbf{a}  - \varepsilon \ \text{log} \ (\mathbf{K}e^{\mathbf{g}^{(\ell)}/\varepsilon} ); \\  &\mathbf{g}^{(\ell+1)} = \varepsilon \ \text{log} \ \mathbf{b}  - \varepsilon \ \text{log} \ (\mathbf{K}^{\mathsf{T}}e^{\mathbf{f}^{(\ell + 1)}/\varepsilon} ). \label{eq:gvaraible} 
	\end{align}
 A side advantage of the updates in~\eq{decayauto} is that it can iteratively decay the regularization parameter through $\varepsilon/\ell$ (see details in \ref{proximal_sinkhorn}).}
 This means we can perform  an \textbf{automatic decaying} on the regularization temperature, and as $\ell\to\infty$ the solution gradually approaches the optimal sparse transportation plan of the original objective (\ref{eq:formulate_pruningnew}). 

  The  gradient of \eq{decayauto}  with respect to the importance scores $\mathbf{s}$ can be obtaind using automatic differentiation.
   We can therefore use normal SGD to jointly train the model weights $\mathbf{w}$ and importance scores $\mathbf{s}$ in \eq{approximate}. 
   
   Our method results in fast training and converges well as shown by our experiments in Section \ref{exp}. 
   After training, we derive the sparse architecture based on the soft masks $\mathbf{m} = n\mathbf{P}^{(L)}\cdot[0, 1]^{\mathsf{T}}$ and finetune the model.
   We outline the iterative training procedure   in \alg{algrithorm}.

\section{Experiments}
\label{exp}


\begin{figure*}[t]
\centering
\begin{tabular}{ccc}
     \includegraphics[width=0.43\textwidth]{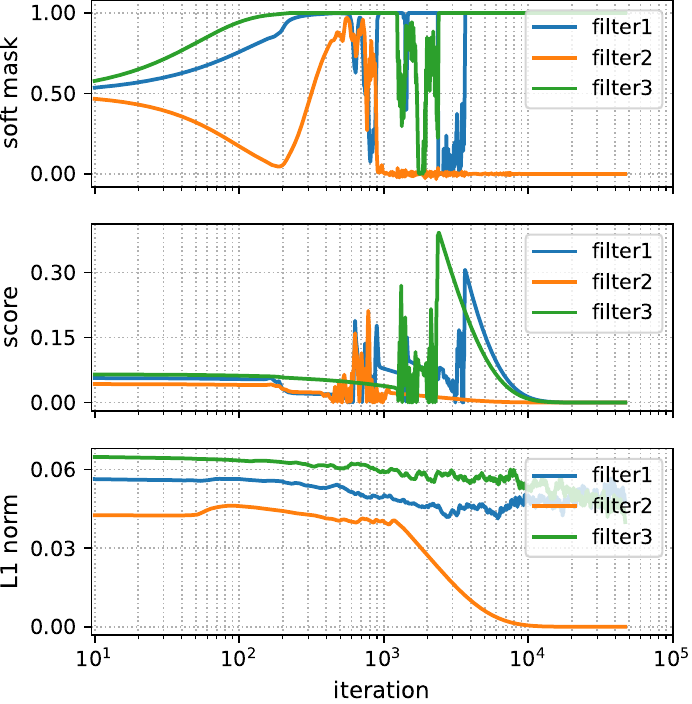} &
         \includegraphics[width=0.542\textwidth]{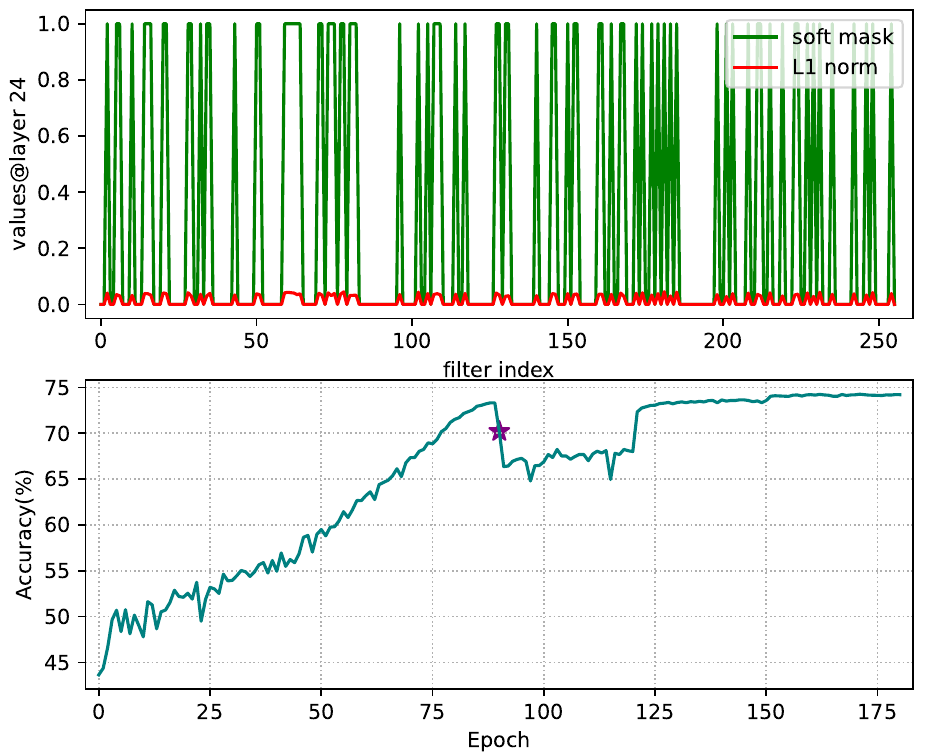}  
         \\  
         \ \ \ \small{(a) 
  On CIFAR-10}  &
  \small{(b) On ImageNet}
    \end{tabular}
    \caption{ 
    Convergence analysis experiments.
   \textit{(a).} 
   Evolution  of the soft masks associated with three filters from
    ResNet-56 trained on CIFAR-10 with $\varepsilon = 1$.   
 Early in training our method explores different filter compositions.  In later training phases the soft masks have converged to hard
masks. For each mask, the soft mask, importance score and L1 norm of the respective filter are correlated. 
  \textit{(b).} 
    Experiments with ResNet-50 on ImageNet with $\varepsilon = 0.25$.  
  After training the soft mask has converged to a discrete hard mask solution. 
The L1 norms of the filters are highly correlated with the soft masks. 
Bottom: Validation accuracy during training with differential transportation pruning and during finetuning. 
The star point shows  test accuracy for derived pruned architecture by soft mask before finetuning. 
    }
 \label{fig:soft_evo}
\end{figure*}

\begin{table*}[t]
\centering
\small
\parbox{.67\linewidth}{{
\centering
\caption{Various pruning ratios with ResNet-56/VGG-19 on CIFAR-10/CIFAR-100:  
	Accuracy comparison with  GReg-1 \cite{wang2021neural} and $L_1$+one-shot~\cite{li2017pruning}.
  Baseline accuracy for ResNet-56 is 93.36\% and for VGG-19 is 74.02\%.
  Each experiment is repeated 3 times with random initialization, mean and std accuracies are reported.
    }
	\renewcommand{\arraystretch}{1.6}
	\resizebox{1\linewidth}{!}{
	 \setlength{\tabcolsep}{0.005mm}{
	   	\begin{tabular}{llccccc}
    	\toprule
		\multirow{5}{*}{\begin{tabular}[c]{@{}c@{}}ResNet-56\\(CIFAR-10)\end{tabular}} & 	 Pruning ratio (\%) & 50  & 70 &  90  & 92.5 & 95 \\
   		 & Sparsity (\%) / Speedup & 49.82/1.99$\times$  & 70.57/3.59$\times$  &  90.39/11.41$\times$   & 93.43/14.76$\times$  & 95.19/19.31$\times$  \\
      \cline{2-7}
		 & $L_1$+one-shot (Acc.\%)~\cite{li2017pruning} & 92.97$\pm$0.15  & 91.88$\pm$0.09 &  87.34$\pm$0.21 &  87.31$\pm$0.28 &  82.79$\pm$0.22 \\
	 &	 GReg-1 (Acc. \%)~\cite{wang2021neural}  & 93.06$\pm$0.09  & 92.23$\pm$0.21 &  89.49$\pm$0.23 &  \textbf{88.39$\pm$0.15} &  85.97$\pm$0.16 \\ 		
	 &	  Ours (Acc. \%) & \textbf{93.46$\pm$0.18}
    &  \textbf{92.46$\pm$0.10}
    &   \textbf{89.84$\pm$0.14}
    &   {88.03$\pm$0.33}
    & \textbf{86.18$\pm$0.06} \\ 		
     \midrule	
    \multirow{5}{*}{\begin{tabular}[c]{@{}c@{}}VGG-19\\(CIFAR-100)\end{tabular}}  &	Pruning ratio (\%) & 50  & 60 &  70  & 80 & 90 \\
    &		 Sparsity (\%) / Speedup & 74.87/3.60$\times$  & 84.00/5.41$\times$  &  90.98/8.84$\times$   & 95.95/17.30$\times$  & 98.96/44.22$\times$  \\
       \cline{2-7}
	 &	 $L_1$+one-shot (Acc.\%)~\cite{li2017pruning} & 71.49$\pm$0.14  & 70.27$\pm$0.12 &  66.05$\pm$0.04 &  61.59$\pm$0.03 &  51.36$\pm$0.11 \\
	 &	 GReg-1 (Acc.\%)~\cite{wang2021neural}  & 71.50$\pm$0.12  & 70.33$\pm$0.12 &  67.35$\pm$0.15 &  63.55$\pm$0.29 &  57.09$\pm$0.03 \\ 		
 &		 Ours (Acc. \%) & \textbf{71.56$\pm$0.19}
    & \textbf{70.41$\pm$0.06}
    &   \textbf{67.74$\pm$0.13}
    &   \textbf{63.98$\pm$0.25}
    &   \textbf{57.21$\pm$0.09}
    \\	
	   \bottomrule
	\end{tabular}
	}}
	\label{tab:resnet56_vgg_cifar}
}}
\hfill
\begin{minipage}{.31\linewidth}
\parbox{1\linewidth}{{
\centering
 	\caption{\RVC{
	Unstructured pruning on ImageNet for ResNet-50.
    }}
	    \renewcommand{\arraystretch}{1}
		\setlength{\tabcolsep}{.5mm}{
	    \resizebox{1\linewidth}{!}{
		\begin{tabular}[c]{lcccccccccc}
		\toprule			
        \multirow{2}*{Method}  &  
        \multirow{2}{*}{\begin{tabular}[c]{@{}c@{}}Baseline\\Acc.(\%)\end{tabular}}
        &   
                \multirow{2}{*}{\begin{tabular}[c]{@{}c@{}}Pruned\\Acc.(\%)\end{tabular}}
        & 
                        \multirow{2}{*}{\begin{tabular}[c]{@{}c@{}}Droped\\ Acc.(\%)\end{tabular}}
        & 
    \multirow{2}{*}{\begin{tabular}[c]{@{}c@{}}Sparsity\\ (\%)\end{tabular}}
        \\
        \\\midrule
   GSM~\cite{ding2019global} & 
   75.72  &  74.30&   1.42 &  80.00 \\ %
      Sparse VD~\cite{molchanov2017variational} 
      & 76.69 &    75.28 &  1.41 & 80.00   \\	
DPF~\cite{lin2020dynamic} &  
75.95 &    74.55 &  1.40 &  82.60  \\	
WoodFisher~\cite{singh2020woodfisher} & 
75.98 &   75.20 &  0.78 &  82.70  \\	
       GReg-1~\cite{wang2021neural} 
       &  76.13 &  75.45 &  0.68 & 82.70           \\	
              GReg-2~\cite{wang2021neural} 
              &  76.13  &    75.27 &  0.86 & 82.70        \\	
              Ours 
              &   76.13  & \textbf{75.50}  &  \textbf{0.63} & 82.70   \\	       
        \bottomrule
		\end{tabular}
  }
  }
\label{tab:imagnet_unstru}
}}
\centering
\vspace{0.01in}
\caption{ \RVC{MobileNetV2 on CIFAR-10/100 with global sparsity ratio over network. }}
\parbox{1\linewidth}{{
\centering
	    \renewcommand{\arraystretch}{0.95}
		\setlength{\tabcolsep}{1.6mm}{
	\resizebox{1\linewidth}{!}{
		\begin{tabular}[c]{lc|ccccccc}
			\toprule
			\multirow{2}{*}{Method} &  \multirow{2}{*}{Sparsity (\%)} & \multicolumn{2}{c}{Acc.} \\
			& & CIFAR-10 & CIFAR-100   \\
			\midrule
   	Unpruned  & 0 &  93.70 & 72.80
   \\
			\midrule

   			ChipNet~\cite{tiwari2021chipnet}  &\multirow{2}{*}{60}& 91.83 & 66.61    \\
			Ours && \textbf{91.93} &  \textbf{70.45}     \\
   \midrule
   			ChipNet~\cite{tiwari2021chipnet}  &\multirow{2}{*}{80}& 90.41   & 52.96    \\
			Ours &&  \textbf{91.49} &  \textbf{68.90}     \\
			\bottomrule
		\end{tabular}
	} \label{tab:gloabalsparsity}	
}	
}}
\end{minipage}
\end{table*}


\RVC{
We validate our proposed method mainly on filter pruning  using different models and datasets, and then extend to unstructured pruning.}
Filter pruning is a form of structured pruning that removes entire filters from the layers of a network, thereby compressing the network, and potentially accelerating the network by reducing the number of FLOPs needed to execute it.

\noindent \textbf{Datasets and networks.}
We first conduct our experiments on CIFAR-10 with ResNet-56 and on CIFAR-100 with VGG-19. 
Following~\cite{wang2021neural}, for the CIFAR datasets we train our baseline models with accuracy comparable to those in the original papers. 
We then evaluate our method on the large-scale ImageNet dataset with ResNet-34 and ResNet-50. Following~\cite{wang2021neural}, we use the PyTorch~\cite{paszke2019pytorch} pretrained models as our baseline models. 
\RVC{
We also apply our method to prune lightweight architecture MobileNetV2. 
}

\noindent \textbf{Training settings.}
The importance scores corresponding to each layer's filters are initialized to the L2-norm of the respective pretrained filter weights. For fair comparisons, we adopt the same pruning rate schedules as~\cite{wang2021neural}.
Also following~\cite{wang2021neural}, for each pruning schedule we denote the model speedup as the pruning-induced reduction in the number of floating point operations (FLOPs) relative to the original model.
The whole training procedure consists of two stages: optimal transportation pruning and post-pruning finetuning. 

Our main focus is on how to train the soft masks in  first stage. 
An important hyperparameter in this respect is  the regularization constant $\varepsilon$.
For ResNet-56 and MobileNetV2, we use $\varepsilon = 1$,  for all other networks we set  $\varepsilon = 0.25$ (see  \ref{Training_details}). 
 The model variables and importance scores are  trained with standard Stochastic Gradient Descent (SGD) with a momentum of $0.9$. For CIFAR datasets, a weight decay of $5\times 10^{-4}$ and a batch size of $256$ are used. 
On ImageNet we use a weight decay of $10^{-4}$ and train on 4 Tesla V100 GPUs with a batch size of $64$ per GPU. 
In all experiments except for MobileNetV2 that follows the training settings of~\cite{tiwari2021chipnet}, the initial learning rate is set as $0.1$ and  a cosine learning rate decay is applied. 
In finetuning stage, we employed the same training settings as~\cite{wang2021neural}.

\subsection{Effect of training soft masks}

\noindent \textbf{How do soft masks converge?}
We show how our soft masks converge to discrete masks during training in this section. 
In \fig{soft_evo}(a), we trained ResNet-56 on  CIFAR-10 with a regularization hyperparameter $\varepsilon = 1$ and a  pruning rate  of 0.5, and visualize the evolution during training of the soft mask, importance score and L1 norm values for three filters. 
Early in training, all filters have similar L1 norms and soft masks. Subsequently, however, the algorithm starts to explore different filter compositions:
The soft mask of filter1 first converges to one, then fluctuates between zero and one, after which it moves back towards one, its final value. The soft masks corresponding to filter2 and filter3 display similar exploratory behavior.  
In later training phases, the soft masks converge to hard masks ($\approx$0 or $\approx$1), and the L1 norm of filters with mask values near zero have gradually decreased to zero, a result of the weight decay term in the loss. Since the mask encodes the neurons' connection information, one could interpret the exploration-exploitation process as performing architecture search.

\RVC{\noindent \textbf{Training overhead with vs without pruning masks.} We measure the wall-clock time for a ResNet-50 on a single A100 GPU for 300 iterations using a batch size of 64 with or without training a pruning mask.
The training overhead increases by $\sim$0.5\%, which is generally negligible.}

\subsection{Accuracy vs. number of Sinkhorn steps}
To improve the computational efficiency of our method, we propose an optimal transportation algorithm that requires only a single Sinkhorn iteration to converge. To verify that a single Sinkhorn iteration indeed suffices, we performed experiments on CIFAR-10 with ResNet-56 using a single GPU. In \fig{sinkhorn_iterations} we plot the accuracy and training time as a function of the number of innner Sinkhorn steps. A larger number of Sinkhorn steps significantly increases training time without markedly affecting accuracy. Thus a single Sinkhorn iteration suffices.


\textbf{\begin{figure}
    \centering
    \includegraphics[width=0.96\linewidth]{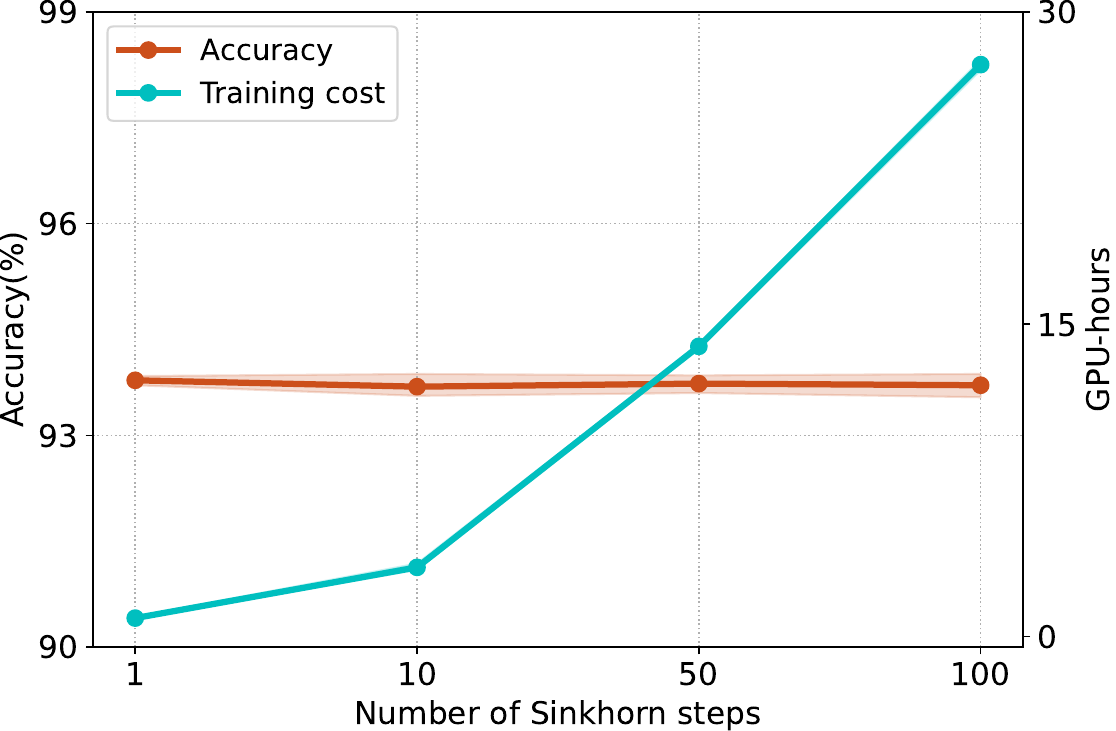}
    \caption{Training cost as a function of the number of inner Sinkhorn steps on CIFAR-10 with ResNet-56 using a single GPU. Each experiment is repeated 3 times with random initialization.
    The training cost for a Sinkhorn step is measured in GPU-hours. 
    Increasing the number of Sinkhorn steps, significantly increases the training cost, but hardly affects the accuracy. 
    A single Sinkhorn step is sufficient for our method.
    }
    \label{fig:sinkhorn_iterations}
\end{figure} }


\subsection{Results for ResNet-56/VGG-19 on CIFAR}

We first explore the effect of applying different pruning schedules on the accuracy of ResNet-56 on CIFAR-10 and VGG-19 on CIFAR-100.
We compare our method to two other pruning methods: L1-norm based one-shot pruning~\cite{li2017pruning} and growing regularization "GReg-1"~\cite{wang2021neural} where the results are taken from the  literature~\cite{wang2021neural}.
We use a simple uniform pruning ratio
scheme, where the pruning ratio is identical across all pruned layers. Following common practice~\cite{gale2019state}, we do not prune the first layer, and for ResNet-56 we only prune the first layer in a residual block (as a result of the restriction imposed by the residual addition) ~\cite{wang2021neural, li2017pruning}. 
We explore a broad range of pruning ratios, covering Speedups between
 $2\times$ and $44\times$.
For fair comparisons, the finetuning scheme (\eg, learning rate schedule, number of epochs, etc.) is identical across all methods.

The results are presented in \tab{resnet56_vgg_cifar}. Note that all pruning methods result in exactly the same pruned network size.
Therefore different observed accuracies between methods result from \emph{how} the filters are pruned.
Our proposed method first performs exploration followed by exploitation. In doing so, our method searches for  best architecture while gradually removing the effect of pruned filters on the network output. Our method 
outperforms L1+one-shot in all examined conditions and GReg-1 for all conditions except one.

\begin{table}[t]
	\centering
 	\caption{
	{Acceleration comparison for structured filter pruning on ImageNet}. 
 Speedup reflects the reduction on FLOPs. 
    }
	    \renewcommand{\arraystretch}{1.1}
		\setlength{\tabcolsep}{1.4mm}{
	    \resizebox{1\linewidth}{!}{
		\begin{tabular}[c]{lccccclccccc}
		\toprule			
        \multirow{2}*{Method}  &  
        \multirow{2}*{Backbone} & 
        \multirow{2}{*}{\begin{tabular}[c]{@{}c@{}}Baseline\\Acc (\%)\end{tabular}}
        &   
                \multirow{2}{*}{\begin{tabular}[c]{@{}c@{}}Pruned\\Acc (\%)\end{tabular}}
        & 
                        \multirow{2}{*}{\begin{tabular}[c]{@{}c@{}}Acc\\Drop (\%)\end{tabular}}
        & 
    \multirow{2}{*}{\begin{tabular}[c]{@{}c@{}}Speed\\ Up\end{tabular}}
        \\
        \\\midrule
   L1 (pruned-B)  ~\cite{li2017pruning} &  \multirow{5}*{ResNet-34} & 73.23  &  72.17&   1.06 &  1.32$\times$  \\ %
      Taylor-FO~\cite{molchanov2019importance} &  & 73.31 &   72.83 &  0.48 & 1.29$\times$   \\	
       GReg-1~\cite{wang2021neural} & & 73.31 &   73.54 &  -0.23 & 1.32$\times$            \\	
              GReg-2~\cite{wang2021neural} & &  73.31  &   73.61 &  -0.30 & 1.32$\times$           \\	
              Ours & &  73.31  & \textbf{74.28}  &  \textbf{-0.97} & 1.32$\times$   \\	       \midrule	
    	 SFP~\cite{he2018soft} & \multirow{13}*{ResNet-50}
      & 76.15 &   74.61 &  1.54 &  1.72$\times$     \\	
	 HRank~\cite{lin2020hrank}&    &    76.15 &  74.98 &  1.17 & 1.78$\times$  \\ 	
		Factorized~\cite{li2019compressing}&  & 76.15 & 74.55 & 1.60 & {2.33$\times$}  \\
				 DCP~\cite{zhuang2018discrimination}&  &  76.01 &  74.95 &  1.06   & 2.25$\times$   \\
		 CCP-AC~\cite{peng2019collaborative} &  &  76.15 &   75.32 & 0.83 & 2.18$\times$  \\
   SRR~\cite{wang2021convolutional}
   &  & 76.13 &  75.11 &  1.02 & 2.27$\times$   \\	         AutoPruner~\cite{luo2020autopruner} & & 76.15 &  74.76 &  1.39 & 2.05$\times$   \\
MetaPruning~\cite{liu2019metapruning} &  & 76.60 &  75.40 &  1.20 & 2.05$\times$   \\	
PGMPF~\cite{cai2022prior} & & 76.01&75.11 & 0.90 & 2.20$\times$  \\
         Random Prune~\cite{li2022revisiting} & & 76.13 &  75.13 &  1.00 & 2.04$\times$   \\
		 GReg-1~\cite{wang2021neural} &  & 76.13 &   75.16 &  0.97 & 2.31$\times$   \\	
       GReg-2~\cite{wang2021neural} &  & 76.13 &   75.36 &  0.77 & 2.31$\times$            \\	
      Ours &  & 76.13 &  \textbf{75.55} &  \textbf{0.58} & 2.31$\times$   \\	
	    \midrule
     		 LFPC~\cite{he2020learning}&  \multirow{4}*{ResNet-50} & 76.15  &  74.46 &   1.69 &  2.55$\times$  \\ %
      GReg-1~\cite{wang2021neural} &  & 76.13 &   74.85 &  1.28 & 2.56$\times$   \\	
       GReg-2~\cite{wang2021neural} &  & 76.13 &   74.93 &  1.20 & 2.56$\times$            \\	
              Ours &   & 76.13 &  \textbf{75.24} &  \textbf{0.89}  & 2.56$\times$   \\	
        \midrule
     		 IncReg~\cite{wang2019structured} &  \multirow{5}*{ResNet-50} & 75.60  &  71.07 &   4.53 &  3.00$\times$  \\ %
      Taylor-FO \cite{molchanov2019importance}&  & 76.18 &   71.69 &  4.49 & 3.05$\times$   \\	
       GReg-1~\cite{wang2021neural} &  & 76.13 &   73.75 &  2.38 & 3.06$\times$            \\	
              GReg-2~\cite{wang2021neural} 
              &  & 76.13 &   73.90 &  2.23 & 3.06$\times$            \\	
              Ours 
              &  & 76.13 &   \textbf{74.26} &  \textbf{1.87} & 3.06$\times$   \\	
        \bottomrule
		\end{tabular}
  }
  }
\label{tab:imagnet}
\vspace{-0.1in}
\end{table}

\subsection{Results for ResNet-34/50 on ImageNet}



We next evaluate our method with ResNet-34 and ResNet-50 on the large-scale ImageNet dataset.
We compare our results to previous pruning methods, including 
L1 Norm~\cite{li2017pruning}, TaylorFO~\cite{molchanov2019importance}, IncReg~\cite{wang2019structured}, SFP~\cite{he2018soft}, HRank~\cite{lin2020hrank}, Factorized~\cite{li2019compressing}, DCP~\cite{zhuang2018discrimination}, CCP-AC~\cite{peng2019collaborative}, LFPC~\cite{he2020learning}, GReg-1 and  GReg-2~\cite{wang2021neural},
SRR~\cite{wang2021convolutional},
MetaPruning~\cite{liu2019metapruning},
AutoPruner~\cite{luo2020autopruner},
PGMPF~\cite{cai2022prior} and
Random Pruning~\cite{li2022revisiting}.
For fair comparisons,
following~\cite{wang2021neural}, we use the official PyTorch ImageNet training example to assure that implementation details such as data augmentation, weight decay, momentum, etc.  match between methods.

We first explore the behavior of our method by training ResNet-50 on ImageNet with the pruning setting corresponding to 3.06× speedup~\cite{wang2021neural} (see detail in \ref{Training_details}). 
In the top row of \fig{soft_evo}(b), we visualize the soft mask and L1 norm of 256 filters from layer 24 of the trained model. Note how all soft masks have converged to hard masks  and   that the L1 norms of the filters are  highly correlated with their corresponding mask values. 
In the bottom row of \fig{soft_evo}(b), we plot the validation accuracy of ResNet-50 during training.
We first use our method to prune the model for 90 epochs and, following previous work~\cite{wang2021neural},  finetune the pruned model for another 90 epochs. 
During the first stage, our method searches for the best architecture and reduces the importance of the pruned filters. After 90 epochs we can safely prune the filters without a large impact on the accuracy, in the star point we  test the accuracy for the derived sparse model before finetuning, after which finetuning  helps  to further improve the accuracy.
At the beginning of finetuning we observe a small accuracy drop since we use a larger learning rate to start with.

\tab{imagnet} gives  an overview of all comparisons. 
We group the methods with similar speedup together for easy comparison. 
We observe that our method consistently achieves
the best result among all approaches, for various speedup comparisons on different architectures.
On ResNet-34, our method improves the baseline (unpruned) model by 0.97\% accuracy.
Previous work~\cite{hoefler2021sparsity} explained how pruning
can improve generalization performance to improve accuracy  for unseen data drawn from the same distribution.
Such improved accuracy after pruning has been observed in earlier work \cite{he2017channel}, but has been more apparent for smaller datasets like CIFAR. In agreement with~\cite{wang2021neural} we also observe improvements over the baseline accuracy on the much more
challenging ImageNet benchmark. 
With larger speedups, the advantage of our method becomes more obvious. For example, our method outperforms
Taylor-FO~\cite{molchanov2019importance}  by 1.05\% top-1 accuracy at the $2.31\times$ setting, while at $3.06\times$,
ours is better by 2.57\% top-1 accuracy.
Previous work has made considerable progress defining importance scores based on the networks weights ~\cite{li2017pruning} ~\cite{he2020learning}. In contrast to those studies our method learns the importance scores used to prune the network in an end-to-end fashion. Our results indicate this can improve the accuracy of pruned networks.

\RVC{
\noindent \textbf{Unstructured pruning on ImageNet.}
Thus far we focused on  structured filter pruning, but previous studies have also explored unstructured pruning. In \tab{imagnet_unstru}, we show that our method can also be applied to unstructured pruning use cases. 
Our method compares favorably to other advanced unstructured pruning methods for ResNet-50 on ImageNet.

\noindent \textbf{Global sparsity for MobileNetV2.}
Thus far we have focused on normal convolutions, but networks based on depthwise convolutions may be harder to prune. In~\tab{gloabalsparsity}, we compare to ChipNet~\cite{tiwari2021chipnet} using the global sparsity ratio for the lightweight  MobileNetV2 architecture on CIFAR-10/100.
Learning with a global sparsity ratio may distribute different sparsity ratios across layers, while respecting the preset sparsity across the network.
For both datasets, we rerun the experiments of ChipNet using 
its released hyperparameters and training settings.
Under different sparsity ratios, our method consistently outperforms ChipNet.

We also further extend our method to general budgeted pruning with \eg a certain FLOPs budget (see detail in \ref{flopslatency}).

}
\section{Conclusion}
\label{conclusion}

In this paper we propose a differentiable sparse learning algorithm based on entropy regularized optimal transportation that achieves exact control over the sparsity ratio. 
We formulate a differentiable
bi-level learning objective to jointly optimize the soft masks
and the network parameters using standard SGD methods combined with Sinkhorn optimization. 
Optimizing this bi-level problem is computational intensive, but we show that by using a Bregman divergence we can improve the efficiency, \ie requiring only a single Sinkhorn iteration,  while still  guaranteeing algorithmic convergence. We demonstrated the effectiveness of our method on different datasets for different pruning rates.



\noindent \textbf{Acknowledgements.} This work is funded by EU’s Horizon Europe research and innovation program under grant agreement No. 101070374.

{\small
\bibliographystyle{ieee_fullname}
\bibliography{main}

\begin{thebibliography}{10}\itemsep=-1pt

\bibitem{afriat1971theory}
SN Afriat.
\newblock Theory of maxima and the method of lagrange.
\newblock {\em SIAM Journal on Applied Mathematics}, 20(3):343--357, 1971.

\bibitem{arjovsky2017wasserstein}
Martin Arjovsky, Soumith Chintala, and L{\'e}on Bottou.
\newblock Wasserstein generative adversarial networks.
\newblock In {\em International conference on machine learning}, pages
  214--223. PMLR, 2017.

\bibitem{cai2022prior}
Linhang Cai, Zhulin An, Chuanguang Yang, Yangchun Yan, and Yongjun Xu.
\newblock Prior gradient mask guided pruning-aware fine-tuning.
\newblock In {\em Proceedings of the AAAI Conference on Artificial
  Intelligence}, volume~1, 2022.

\bibitem{censor1992proximal}
Yair Censor and Stavros~Andrea Zenios.
\newblock Proximal minimization algorithm with d-functions.
\newblock {\em Journal of Optimization Theory and Applications},
  73(3):451--464, 1992.

\bibitem{changpinyo2017power}
Soravit Changpinyo, Mark Sandler, and Andrey Zhmoginov.
\newblock The power of sparsity in convolutional neural networks.
\newblock {\em arXiv preprint arXiv:1702.06257}, 2017.

\bibitem{chijiwa2021pruning}
Daiki Chijiwa, Shin'ya Yamaguchi, Yasutoshi Ida, Kenji Umakoshi, and Tomohiro
  Inoue.
\newblock Pruning randomly initialized neural networks with iterative
  randomization.
\newblock {\em Advances in Neural Information Processing Systems},
  34:4503--4513, 2021.

\bibitem{colson2007overview}
Beno{\^\i}t Colson, Patrice Marcotte, and Gilles Savard.
\newblock An overview of bilevel optimization.
\newblock {\em Annals of operations research}, 153(1):235--256, 2007.

\bibitem{cuturi2013sinkhorn}
Marco Cuturi.
\newblock Sinkhorn distances: Lightspeed computation of optimal transport.
\newblock {\em Advances in neural information processing systems}, 26, 2013.

\bibitem{ding2019global}
Xiaohan Ding, Xiangxin Zhou, Yuchen Guo, Jungong Han, Ji Liu, et~al.
\newblock Global sparse momentum sgd for pruning very deep neural networks.
\newblock {\em Advances in Neural Information Processing Systems}, 32, 2019.

\bibitem{ditschuneit2022auto}
Konstantin Ditschuneit and Johannes~S Otterbach.
\newblock Auto-compressing subset pruning for semantic image segmentation.
\newblock In {\em Pattern Recognition: 44th DAGM German Conference}, pages
  20--35. Springer, 2022.

\bibitem{eisenberger2022unified}
Marvin Eisenberger, Aysim Toker, Laura Leal-Taix{\'e}, Florian Bernard, and
  Daniel Cremers.
\newblock A unified framework for implicit sinkhorn differentiation.
\newblock In {\em Proceedings of the IEEE/CVF Conference on Computer Vision and
  Pattern Recognition}, pages 509--518, 2022.

\bibitem{finn2017model}
Chelsea Finn, Pieter Abbeel, and Sergey Levine.
\newblock Model-agnostic meta-learning for fast adaptation of deep networks.
\newblock In {\em International conference on machine learning}, pages
  1126--1135. PMLR, 2017.

\bibitem{frankle2018lottery}
Jonathan Frankle and Michael Carbin.
\newblock The lottery ticket hypothesis: Finding sparse, trainable neural
  networks.
\newblock {\em ICLR}, 2019.

\bibitem{gale2019state}
Trevor Gale, Erich Elsen, and Sara Hooker.
\newblock The state of sparsity in deep neural networks.
\newblock {\em arXiv preprint arXiv:1902.09574}, 2019.

\bibitem{gao2020discrete}
Shangqian Gao, Feihu Huang, Jian Pei, and Heng Huang.
\newblock Discrete model compression with resource constraint for deep neural
  networks.
\newblock In {\em Proceedings of the IEEE/CVF Conference on Computer Vision and
  Pattern Recognition}, pages 1899--1908, 2020.

\bibitem{genevay2018learning}
Aude Genevay, Gabriel Peyr{\'e}, and Marco Cuturi.
\newblock Learning generative models with sinkhorn divergences.
\newblock In {\em International Conference on Artificial Intelligence and
  Statistics}, pages 1608--1617. PMLR, 2018.

\bibitem{guo2020dmcp}
Shaopeng Guo, Yujie Wang, Quanquan Li, and Junjie Yan.
\newblock Dmcp: Differentiable markov channel pruning for neural networks.
\newblock In {\em Proceedings of the IEEE/CVF conference on computer vision and
  pattern recognition}, pages 1539--1547, 2020.

\bibitem{he2020learning}
Yang He, Yuhang Ding, Ping Liu, Linchao Zhu, Hanwang Zhang, and Yi Yang.
\newblock Learning filter pruning criteria for deep convolutional neural
  networks acceleration.
\newblock In {\em Proceedings of the IEEE/CVF conference on computer vision and
  pattern recognition}, pages 2009--2018, 2020.

\bibitem{he2018soft}
Yang He, Guoliang Kang, Xuanyi Dong, Yanwei Fu, and Yi Yang.
\newblock Soft filter pruning for accelerating deep convolutional neural
  networks.
\newblock {\em IJCAI}, 2018.

\bibitem{he2019filter}
Yang He, Ping Liu, Ziwei Wang, Zhilan Hu, and Yi Yang.
\newblock Filter pruning via geometric median for deep convolutional neural
  networks acceleration.
\newblock In {\em Proceedings of the IEEE/CVF conference on computer vision and
  pattern recognition}, pages 4340--4349, 2019.

\bibitem{he2017channel}
Yihui He, Xiangyu Zhang, and Jian Sun.
\newblock Channel pruning for accelerating very deep neural networks.
\newblock In {\em Proceedings of the IEEE international conference on computer
  vision}, pages 1389--1397, 2017.

\bibitem{hoefler2021sparsity}
Torsten Hoefler, Dan Alistarh, Tal Ben-Nun, Nikoli Dryden, and Alexandra Peste.
\newblock Sparsity in deep learning: Pruning and growth for efficient inference
  and training in neural networks.
\newblock {\em J. Mach. Learn. Res.}, 22(241):1--124, 2021.

\bibitem{huang2018data}
Zehao Huang and Naiyan Wang.
\newblock Data-driven sparse structure selection for deep neural networks.
\newblock In {\em Proceedings of the European conference on computer vision
  (ECCV)}, pages 304--320, 2018.

\bibitem{Humble2022pruning}
Ryan Humble, Maying Shen, Jorge Albericio-Latorre, Eric Darve, and Jose~M
  Alvarez.
\newblock Soft masking for cost-constrained channel pruning.
\newblock {\em ECCV}, 2022.

\bibitem{kang2020operation}
Minsoo Kang and Bohyung Han.
\newblock Operation-aware soft channel pruning using differentiable masks.
\newblock In {\em International Conference on Machine Learning}, pages
  5122--5131. PMLR, 2020.

\bibitem{kosowsky1994invisible}
Jeffrey~J Kosowsky and Alan~L Yuille.
\newblock The invisible hand algorithm: Solving the assignment problem with
  statistical physics.
\newblock {\em Neural networks}, 7(3):477--490, 1994.

\bibitem{krashinsky2020nvidia}
Ronny Krashinsky, Olivier Giroux, Stephen Jones, Nick Stam, and Sridhar
  Ramaswamy.
\newblock Nvidia ampere architecture in-depth.
\newblock {\em NVIDIA blog: https://devblogs. nvidia.
  com/nvidia-ampere-architecture-in-depth}, 2020.

\bibitem{lemaire2019structured}
Carl Lemaire, Andrew Achkar, and Pierre-Marc Jodoin.
\newblock Structured pruning of neural networks with budget-aware
  regularization.
\newblock In {\em Proceedings of the IEEE/CVF Conference on Computer Vision and
  Pattern Recognition}, pages 9108--9116, 2019.

\bibitem{li2020eagleeye}
Bailin Li, Bowen Wu, Jiang Su, and Guangrun Wang.
\newblock Eagleeye: Fast sub-net evaluation for efficient neural network
  pruning.
\newblock In {\em ECCV 2020}, pages 639--654. Springer, 2020.

\bibitem{li2017pruning}
Hao Li, Asim Kadav, Igor Durdanovic, Hanan Samet, and Hans~Peter Graf.
\newblock Pruning filters for efficient convnets.
\newblock {\em ICLR}, 2017.

\bibitem{li2019compressing}
Tuanhui Li, Baoyuan Wu, Yujiu Yang, Yanbo Fan, Yong Zhang, and Wei Liu.
\newblock Compressing convolutional neural networks via factorized
  convolutional filters.
\newblock In {\em Proceedings of the IEEE/CVF Conference on Computer Vision and
  Pattern Recognition}, pages 3977--3986, 2019.

\bibitem{li2022revisiting}
Yawei Li, Kamil Adamczewski, Wen Li, Shuhang Gu, Radu Timofte, and Luc
  Van~Gool.
\newblock Revisiting random channel pruning for neural network compression.
\newblock In {\em Proceedings of the IEEE/CVF Conference on Computer Vision and
  Pattern Recognition}, pages 191--201, 2022.

\bibitem{liebenwein2019provable}
Lucas Liebenwein, Cenk Baykal, Harry Lang, Dan Feldman, and Daniela Rus.
\newblock Provable filter pruning for efficient neural networks.
\newblock {\em arXiv preprint arXiv:1911.07412}, 2019.

\bibitem{lin2020hrank}
Mingbao Lin, Rongrong Ji, Yan Wang, Yichen Zhang, Baochang Zhang, Yonghong
  Tian, and Ling Shao.
\newblock Hrank: Filter pruning using high-rank feature map.
\newblock In {\em Proceedings of the IEEE/CVF conference on computer vision and
  pattern recognition}, pages 1529--1538, 2020.

\bibitem{lin2020dynamic}
Tao Lin, Sebastian~U Stich, Luis Barba, Daniil Dmitriev, and Martin Jaggi.
\newblock Dynamic model pruning with feedback.
\newblock {\em arXiv preprint arXiv:2006.07253}, 2020.

\bibitem{liu2019darts}
Hanxiao Liu, Karen Simonyan, and Yiming Yang.
\newblock Darts: Differentiable architecture search.
\newblock {\em ICLR}, 2019.

\bibitem{liu2019metapruning}
Zechun Liu, Haoyuan Mu, Xiangyu Zhang, Zichao Guo, Xin Yang, Kwang-Ting Cheng,
  and Jian Sun.
\newblock Metapruning: Meta learning for automatic neural network channel
  pruning.
\newblock In {\em Proceedings of the IEEE/CVF international conference on
  computer vision}, pages 3296--3305, 2019.

\bibitem{liu2018rethinking}
Zhuang Liu, Mingjie Sun, Tinghui Zhou, Gao Huang, and Trevor Darrell.
\newblock Rethinking the value of network pruning.
\newblock {\em ICLR}, 2019.

\bibitem{luketina2016scalable}
Jelena Luketina, Mathias Berglund, Klaus Greff, and Tapani Raiko.
\newblock Scalable gradient-based tuning of continuous regularization
  hyperparameters.
\newblock In {\em International conference on machine learning}, pages
  2952--2960. PMLR, 2016.

\bibitem{luo2020autopruner}
Jian-Hao Luo and Jianxin Wu.
\newblock Autopruner: An end-to-end trainable filter pruning method for
  efficient deep model inference.
\newblock {\em Pattern Recognition}, 107:107461, 2020.

\bibitem{luo2020neural}
Jian-Hao Luo and Jianxin Wu.
\newblock Neural network pruning with residual-connections and limited-data.
\newblock In {\em Proceedings of the IEEE/CVF Conference on Computer Vision and
  Pattern Recognition}, pages 1458--1467, 2020.

\bibitem{meng2020pruning}
Fanxu Meng, Hao Cheng, Ke Li, Huixiang Luo, Xiaowei Guo, Guangming Lu, and Xing
  Sun.
\newblock Pruning filter in filter.
\newblock {\em Advances in Neural Information Processing Systems},
  33:17629--17640, 2020.

\bibitem{miao2021learning}
Lu Miao, Xiaolong Luo, Tianlong Chen, Wuyang Chen, Dong Liu, and Zhangyang
  Wang.
\newblock Learning pruning-friendly networks via frank-wolfe: One-shot,
  any-sparsity, and no retraining.
\newblock In {\em International Conference on Learning Representations}, 2022.

\bibitem{molchanov2017variational}
Dmitry Molchanov, Arsenii Ashukha, and Dmitry Vetrov.
\newblock Variational dropout sparsifies deep neural networks.
\newblock In {\em International Conference on Machine Learning}, pages
  2498--2507. PMLR, 2017.

\bibitem{molchanov2019importance}
Pavlo Molchanov, Arun Mallya, Stephen Tyree, Iuri Frosio, and Jan Kautz.
\newblock Importance estimation for neural network pruning.
\newblock In {\em Proceedings of the IEEE/CVF Conference on Computer Vision and
  Pattern Recognition}, pages 11264--11272, 2019.

\bibitem{ning2020dsa}
Xuefei Ning, Tianchen Zhao, Wenshuo Li, Peng Lei, Yu Wang, and Huazhong Yang.
\newblock Dsa: More efficient budgeted pruning via differentiable sparsity
  allocation.
\newblock In {\em ECCV 2020}, pages 592--607. Springer, 2020.

\bibitem{parikh2014proximal}
Neal Parikh, Stephen Boyd, et~al.
\newblock Proximal algorithms.
\newblock {\em Foundations and trends{\textregistered} in Optimization},
  1(3):127--239, 2014.

\bibitem{paszke2017automatic}
Adam Paszke, Sam Gross, Soumith Chintala, Gregory Chanan, Edward Yang, Zachary
  DeVito, Zeming Lin, Alban Desmaison, Luca Antiga, and Adam Lerer.
\newblock Automatic differentiation in pytorch.
\newblock 2017.

\bibitem{paszke2019pytorch}
Adam Paszke, Sam Gross, Francisco Massa, Adam Lerer, James Bradbury, Gregory
  Chanan, Trevor Killeen, Zeming Lin, Natalia Gimelshein, Luca Antiga, et~al.
\newblock Pytorch: An imperative style, high-performance deep learning library.
\newblock {\em Advances in neural information processing systems}, 32, 2019.

\bibitem{peng2019collaborative}
Hanyu Peng, Jiaxiang Wu, Shifeng Chen, and Junzhou Huang.
\newblock Collaborative channel pruning for deep networks.
\newblock In {\em International Conference on Machine Learning}, pages
  5113--5122. PMLR, 2019.

\bibitem{peyre2019computational}
Gabriel Peyr{\'e}, Marco Cuturi, et~al.
\newblock Computational optimal transport: With applications to data science.
\newblock {\em Foundations and Trends{\textregistered} in Machine Learning},
  11(5-6):355--607, 2019.

\bibitem{savarese2020winning}
Pedro Savarese, Hugo Silva, and Michael Maire.
\newblock Winning the lottery with continuous sparsification.
\newblock {\em Advances in Neural Information Processing Systems},
  33:11380--11390, 2020.

\bibitem{schmitzer2019stabilized}
Bernhard Schmitzer.
\newblock Stabilized sparse scaling algorithms for entropy regularized
  transport problems.
\newblock {\em arXiv preprint arXiv:1610.06519}, 2016.

\bibitem{shen2022structural}
Maying Shen, Hongxu Yin, Pavlo Molchanov, Lei Mao, Jianna Liu, and Jose
  Alvarez.
\newblock Structural pruning via latency-saliency knapsack.
\newblock In {\em Advances in Neural Information Processing Systems}, 2022.

\bibitem{shen2020cpot}
Yucong Shen, Li Shen, Hao-Zhi Huang, Xuan Wang, and Wei Liu.
\newblock Cpot: Channel pruning via optimal transport.
\newblock {\em arXiv preprint arXiv:2005.10451}, 2020.

\bibitem{singh2020woodfisher}
Sidak~Pal Singh and Dan Alistarh.
\newblock Woodfisher: Efficient second-order approximation for neural network
  compression.
\newblock {\em Advances in Neural Information Processing Systems},
  33:18098--18109, 2020.

\bibitem{sutskever2013importance}
Ilya Sutskever, James Martens, George Dahl, and Geoffrey Hinton.
\newblock On the importance of initialization and momentum in deep learning.
\newblock In {\em International conference on machine learning}, pages
  1139--1147. PMLR, 2013.

\bibitem{suzuki2001simple}
Kenji Suzuki, Isao Horiba, and Noboru Sugie.
\newblock A simple neural network pruning algorithm with application to filter
  synthesis.
\newblock {\em Neural processing letters}, 13(1):43--53, 2001.

\bibitem{tai2022spartan}
Kai~Sheng Tai, Taipeng Tian, and Ser~Nam Lim.
\newblock Spartan: Differentiable sparsity via regularized transportation.
\newblock {\em Advances in Neural Information Processing Systems},
  35:4189--4202, 2022.

\bibitem{tiwari2021chipnet}
Rishabh Tiwari, Udbhav Bamba, Arnav Chavan, and Deepak Gupta.
\newblock Chipnet: Budget-aware pruning with heaviside continuous
  approximations.
\newblock In {\em International Conference on Learning Representations}, 2021.

\bibitem{vischer2021lottery}
Marc~Aurel Vischer, Robert~Tjarko Lange, and Henning Sprekeler.
\newblock On lottery tickets and minimal task representations in deep
  reinforcement learning.
\newblock {\em ICLR}, 2022.

\bibitem{wang2019structured}
Huan Wang, Xinyi Hu, Qiming Zhang, Yuehai Wang, Lu Yu, and Haoji Hu.
\newblock Structured pruning for efficient convolutional neural networks via
  incremental regularization.
\newblock {\em IEEE Journal of Selected Topics in Signal Processing},
  14(4):775--788, 2019.

\bibitem{wang2021neural}
Huan Wang, Can Qin, Yulun Zhang, and Yun Fu.
\newblock Neural pruning via growing regularization.
\newblock {\em ICLR}, 2021.

\bibitem{wang2021convolutional}
Zi Wang, Chengcheng Li, and Xiangyang Wang.
\newblock Convolutional neural network pruning with structural redundancy
  reduction.
\newblock In {\em Proceedings of the IEEE/CVF Conference on Computer Vision and
  Pattern Recognition}, pages 14913--14922, 2021.

\bibitem{xie2020differentiable}
Yujia Xie, Hanjun Dai, Minshuo Chen, Bo Dai, Tuo Zhao, Hongyuan Zha, Wei Wei,
  and Tomas Pfister.
\newblock Differentiable top-k with optimal transport.
\newblock {\em Advances in Neural Information Processing Systems},
  33:20520--20531, 2020.

\bibitem{xie2020fast}
Yujia Xie, Xiangfeng Wang, Ruijia Wang, and Hongyuan Zha.
\newblock A fast proximal point method for computing exact wasserstein
  distance.
\newblock In {\em Uncertainty in artificial intelligence}, pages 433--453.
  PMLR, 2020.

\bibitem{ye2018rethinking}
Jianbo Ye, Xin Lu, Zhe Lin, and James~Z Wang.
\newblock Rethinking the smaller-norm-less-informative assumption in channel
  pruning of convolution layers.
\newblock {\em arXiv preprint arXiv:1802.00124}, 2018.

\bibitem{zhang2021lottery}
Shuai Zhang, Meng Wang, Sijia Liu, Pin-Yu Chen, and Jinjun Xiong.
\newblock Why lottery ticket wins? a theoretical perspective of sample
  complexity on sparse neural networks.
\newblock {\em Advances in Neural Information Processing Systems},
  34:2707--2720, 2021.

\bibitem{zhuang2018discrimination}
Zhuangwei Zhuang, Mingkui Tan, Bohan Zhuang, Jing Liu, Yong Guo, Qingyao Wu,
  Junzhou Huang, and Jinhui Zhu.
\newblock Discrimination-aware channel pruning for deep neural networks.
\newblock {\em Advances in neural information processing systems}, 31, 2018.

\end{thebibliography}
}

\clearpage

\appendix

\section{Supplementary
Material}
\label{appendix}
\subsection{Optimal coupling to regularized problem}
\label{Optimal_coupling}
\noindent
Here we prove the optimal  coupling to the regularized problem in \eq{entropyopt2} is:
    \begin{equation}
\mathbf{P}_{\varepsilon} = e ^{{\mathbf{f}}/{\varepsilon}} \odot e ^{-{\mathbf{C}}/{\varepsilon}} \odot e ^{{\mathbf{g}}/{\varepsilon}}.
\label{eq:appen_lagrange}
\end{equation}
\textit{Proof.}
Introducing two dual variables $\mathbf{f}\in \mathbb{R}^{n}$ and $\mathbf{g}\in \mathbb{R}^{2}$ for marginal constraints
$\mathbf{P} 
 \mathbbm{1}_2 = \mathbf{a}$ and $ \mathbf{P}^\mathsf{T} \mathbbm{1}_n = \mathbf{b}$, given the discrete entropy of a coupling matrix  $\mathcal{H}(\mathbf{P}) = -\sum_{ij}\mathbf{P}_{ij}(\text{log}(\mathbf{P}_{ij}) - 1)$,  
the  Lagrangian of \eq{entropyopt2} is optimized as follows:
   \begin{equation}
	{\xi}(\mathbf{P}, \mathbf{f}, \mathbf{g}) = \left \langle \mathbf{C}, \mathbf{P}\right \rangle - \varepsilon \mathcal{H} (\mathbf{P}) - \left \langle \mathbf{f}, \mathbf{P} 
 \mathbbm{1}_2 - \mathbf{a}\right \rangle- \left \langle \mathbf{g}, \mathbf{P}^\mathsf{T} 
 \mathbbm{1}_n - \mathbf{b}\right \rangle.
\label{eq:appen_entropyopt2}
\end{equation}
First order conditions then yield:
   \begin{equation}
	\frac{\partial{\xi}(\mathbf{P}, \mathbf{f}, \mathbf{g})}{\partial \mathbf{P}_{ij}} = \mathbf{C} + \varepsilon \ \text{log}(\mathbf{P}_{ij}) -  \mathbf{f}_{i}- \mathbf{g}_{j} = 0,
\label{eq:appen_Firstorder}
\end{equation}
which result for an optimal coupling to the regularized problem, in the  matrix expression shown in  \eq{appen_lagrange}.

\subsection{Dual problem computation}
\noindent
Here we prove that
minimizing the regularized optimal transport distance in \eq{entropyopt2}  is equivalent to maximizing its dual problem:
\label{Dual_problem}
 \begin{align}
\mathop{\max}_{
\mathbf{f}\in \mathbb{R}^{n}, \mathbf{g}\in \mathbb{R}^{2}
} \left \langle \mathbf{f}, \mathbf{a}\right \rangle
+ \left \langle \mathbf{g}, \mathbf{b}\right \rangle
- \varepsilon \left \langle e ^{{\mathbf{f}}/{\varepsilon}},  e^{-{\mathbf{C}{(\mathbf{s})}}/{\varepsilon}} \cdot e ^{{\mathbf{g}}/{\varepsilon}}\right \rangle
\label{eq:appendix_dual}
\end{align}
\textit{Proof.} We start from the result in \eq{appen_lagrange}, and substitute it in the Lagrangian ${\xi}(\mathbf{P}, \mathbf{f}, \mathbf{g})$ of \eq{appen_entropyopt2}, where the optimal $\mathbf{P}$ is  a function of $\mathbf{f}$ and $\mathbf{g}$, we obtain that the Lagrange dual function equals:
  \begin{equation}
\begin{split}
  \mathbf{f},  \mathbf{g} \mapsto
  & \left \langle e ^{{\mathbf{f}}/{\varepsilon}},  \big(e^{-{\mathbf{C}{(\mathbf{s})}}/{\varepsilon}}  \odot \mathbf{C}{(\mathbf{s})}\big) e ^{{\mathbf{g}}/{\varepsilon}}\right \rangle \\
  & - \varepsilon \mathcal{H} \big(e ^{{\mathbf{f}}/{\varepsilon}} \odot e ^{-{\mathbf{C}{(\mathbf{s})}}/{\varepsilon}} \odot e ^{{\mathbf{g}}/{\varepsilon}} \big) 
\end{split}
\label{eq:appendix_dual1}
\end{equation}
The negative entropy of $\mathbf{P}$ scaled by $\varepsilon$, namely $\varepsilon \left \langle \mathbf{P}, \ \text{log}\mathbf{P} -\mathbbm{1}_{n \times 2} \right \rangle$, can be stated explicitly
as a function of $ \mathbf{f},  \mathbf{g},  \mathbf{C}$:
 \begin{align*}
  &\varepsilon \left \langle 
 \mathbf{P}, \ \text{log}\mathbf{P} -\mathbbm{1}_{n \times 2} \right \rangle \\
 = &
   \left \langle 
e ^{{\mathbf{f}}/{\varepsilon}} \odot e ^{-{\mathbf{C}{(\mathbf{s})}}/{\varepsilon}} \odot e ^{{\mathbf{g}}/{\varepsilon}},  \mathbf{f} \mathbbm{1}_{2}^\mathsf{T} + \mathbbm{1}_{n}\mathbf{g}^\mathsf{T} - \mathbf{C}{(\mathbf{s})} 
-  \varepsilon\mathbbm{1}_{n \times 2}   \right \rangle  \\
 = & - \left \langle e ^{{\mathbf{f}}/{\varepsilon}},  \big(e^{-{\mathbf{C}{(\mathbf{s})}}/{\varepsilon}}  \odot \mathbf{C}{(\mathbf{s})}\big) e ^{{\mathbf{g}}/{\varepsilon}}\right \rangle
+ \left \langle \mathbf{f}, \mathbf{a}\right \rangle
+ \left \langle \mathbf{g}, \mathbf{b}\right \rangle
- \\
& \varepsilon \left \langle e ^{{\mathbf{f}}/{\varepsilon}},  e^{-{\mathbf{C}{(\mathbf{s})}}/{\varepsilon}} \cdot e ^{{\mathbf{g}}/{\varepsilon}}\right \rangle
\label{eq:appendix_dual2}
\end{align*}
therefore, the first term in \eq{appendix_dual1} cancels out with the first term in the entropy above.
The remaining terms are those appearing in \eq{appendix_dual}.

\subsection{Expensive to train with Sinkhorn’s algorithm}
\label{For_back_sinkhorn}
\begin{figure*}
    \centering
\includegraphics[width=0.99\linewidth]{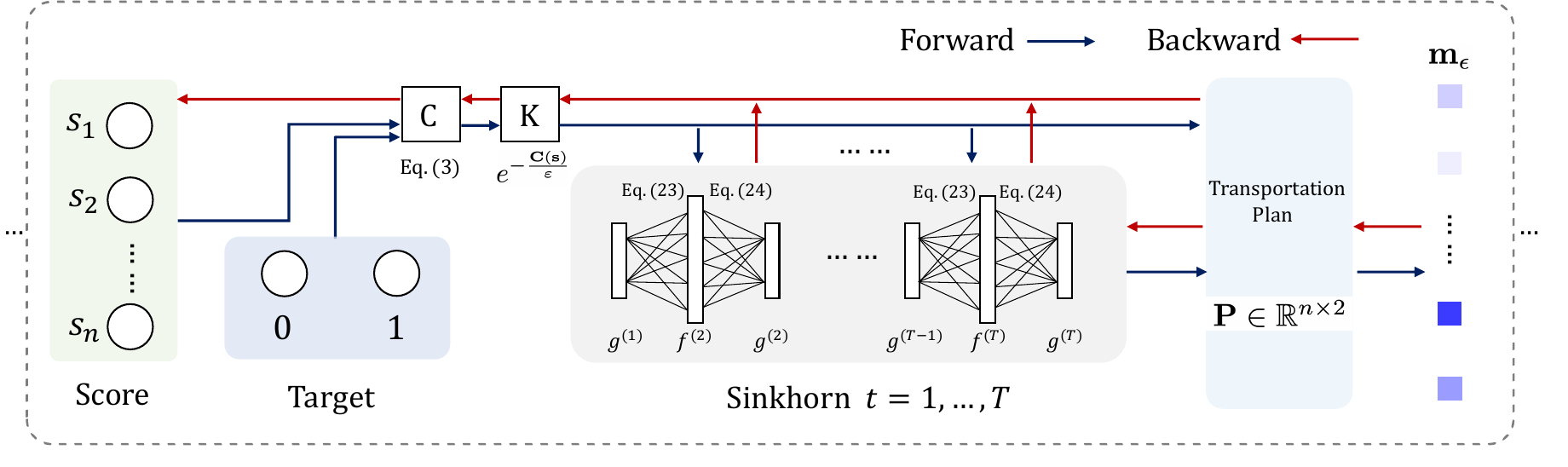}
    \caption{The display of training with Sinkhorn’s algorithm. 
    We show the forward and backward pass over one single layer: given importance score $\mathbf{s}$, the cost matrix is computed by \eq{cost_C},  we thus get $\mathbf{K}$; in forward pass we  input  $\mathbf{K}$ to iteratively compute
    $\mathbf{f}^{(t+1)}$ with \eq{appen_gvaraibleupp} and $\mathbf{g}^{(t+1)}$ with \eq{appen_gvaraible1} using Sinkhorn algorithm for $T$ iterations; during  back-propagation pass, the
gradient of soft mask with respect to the importance scores  is
computed by differentiating through all Sinkhorn iterations.  A large $T$ makes it expensive to train.
    } 
    \label{fig:app_sinkhornoverview}
\end{figure*}

\noindent
The bi-level optimization problem in
\eq{upper}  and \eq{lowerl}
is expensive to train with Sinkhorn's algorithm:
 During forward pass, per mini-batch the inner optimization  in \eq{lowerl} needs to perform Sinkhorn's iterative algorithm for hundreds of iterations to converge, which is computationally inefficient. 
During the back-propagation pass, the gradient of $\mathbf{P}^*_{\varepsilon}(\mathbf{s})$
 with respect to the importance scores  $\mathbf{s}$ is computed by differentiating~\cite{paszke2017automatic} through all Sinkhorn iterations, which is expensive. 
We display the 
 forward and backward of  Sinkhorn algorithm  in \fig{app_sinkhornoverview}.   

\begin{table*}[t]
	\centering
\caption{Training setting:  For the SGD optimizer, in the parentheses are the momentum and
weight decay. For ImageNet, batch size is 64 per GPU. We learn the soft mask using cosine learning rate schedule. 
    }
	\renewcommand{\arraystretch}{1.}
	\resizebox{1\linewidth}{!}{
	 \setlength{\tabcolsep}{10mm}{
	   	\begin{tabular}{l|c|ccc}
    	\toprule
  Settings  & CIFAR & ImageNet  \\
      \midrule	
		  Optimizer &  
    SGD (0.9, 5e-4) & SGD (0.9, 1e-4)    \\
     \midrule	
	 	 LR schedule (soft mask) 
    & \multicolumn{2}{c}{Cosine LR schedule (0.1)} 
    \\ 		
        \midrule
	 	 LR schedule (finetune) & Multi-step (0:1e-2, 60:1e-3, 90:1e-4)  
    & Multi-step (0:1e-2, 30:1e-3, 60:1e-4, 75:1e-5)
    \\ 		
        \midrule
Training Epoch&  120  + 120
    & 90 + 90
    \\ 	
        \midrule
Batch size & 256 
    & 256
    \\ 	
\bottomrule
\end{tabular}
}
}
\label{tab:trainingsetting}
\end{table*}
\subsection{Bregman divergence based optimization}
\label{Bregman}
\noindent
In this paper we use similar Bregman Divergence $D_h$  as 
 \cite{xie2020fast} based on entropy function $\mathcal{H}(\mathbf{x}) = -\sum_{i}{x}_{i} \ \big(\text{log}(x_{i}) - 1\big)$ as:
 \begin{align}
D_{\mathcal{H}}(\mathbf{x}, \mathbf{y}) &= -\sum_{i}x_{i} \ \text{log}\frac{{x}_{i}}{{y}_{i}} + \sum_{i}{x}_{i}.
\end{align}
\label{eq:Bregman_proximal}
Based on the defined Bregman Divergence, a single proximal point iteration for problem (\ref{eq:opt})
can be written as:
   \begin{equation}
\begin{split}
\mathbf{P}^{(\ell+1)} &= \mathop{\min}_{\mathbf{P} \in\mathcal{U}(\mathbf{a}, \mathbf{b})}	\left \langle \mathbf{C}, \mathbf{P}\right \rangle - \varepsilon D_{\mathcal{H}}(\mathbf{\mathbf{P}, \mathbf{P}^{(\ell)}}) \\
&=  \mathop{\min}_{\mathbf{P} \in\mathcal{U}(\mathbf{a}, \mathbf{b})}	\left \langle \mathbf{C} - \varepsilon \ \text{log} \ \mathbf{P}^{(\ell)}, \mathbf{P}\right \rangle - \varepsilon {\mathcal{H}}(\mathbf{\mathbf{P}}).
\end{split}
\label{eq:appentropyopt3}
\end{equation}
Denote $\mathbf{C}^{\prime} = \mathbf{C} - \varepsilon \ \text{log} \ \mathbf{P}^{(\ell)}$. Note that for optimization
problem \ref{eq:appentropyopt3}, $\mathbf{P}^{(\ell)}$
is a fixed value that is not relevant to
optimization variable $\mathbf{P}$. Comparing
to \eq{entropyopt2}, the problem in \eq{appentropyopt3} can be solved by Sinkhorn iteration
by replacing Gibbs kernel $\mathbf{K}$ by $ \mathbf{K}^{\prime} = e^{-\frac{\mathbf{C}^{\prime}}{\varepsilon}} = e^{-\frac{\mathbf{C}}{\varepsilon}} \odot \mathbf{P}^{(\ell)}$. 
With this
reorganization, we can solve it with Sinkhorn algorithm.
\cite{xie2020fast} have shown both theoretically and empirically
that a
\textit{ single
Sinkhorn inner iteration} is  sufficient to converge under a large range of fixed $\varepsilon$, we therefore compute  $\mathbf{P}^{(\ell+1)}$  as:
  \begin{align}
\mathbf{P}^{(\ell+1)} &= e ^{{\mathbf{f}^{(\ell + 1)}}/{\varepsilon}}  \odot \big(e^{-\frac{\mathbf{C}}{\varepsilon}} \odot \mathbf{P}^{(\ell)}\big)  \odot e ^{{\mathbf{g}^{(\ell + 1)}}/{\varepsilon}},
\label{eq:appdecayauto0}
\end{align}
 which is the expression of \eq{decayauto}. 


\subsection{Proximal point iteration as iterative Sinkhorn}
\label{proximal_sinkhorn}
\noindent We  refer to the proof  in Chapter 4.2 of the  book~\cite{peyre2019computational}.
The general setting for proximal point iteration is to define Gibbs kernel as  $\mathbf{K} = e^{-\frac{\mathbf{C}}{\varepsilon}} \odot \mathbf{P}^{(\ell)}$, the proximal point iterations thus have the form:
   \begin{equation}
\begin{split}
\mathbf{P}^{(\ell+1)}_\varepsilon(\mathbf{s}) = & e ^{{\mathbf{f}^{(\ell+ 1)}}/{\varepsilon}}  \odot \big(e^{-\frac{\mathbf{C}}{\varepsilon}} \odot \mathbf{P}^{(\ell)}\big)  \odot e ^{{\mathbf{g}^{(\ell+ 1)}}/{\varepsilon}}  \\
 = &\big(e ^{{\mathbf{f}^{(\ell+ 1)}}/{\varepsilon}} \odot 
\cdot \cdot  \odot e ^{{\mathbf{f}^{(1)}}/{\varepsilon}} 
\big)
\odot \big(e^{-\frac{(\ell + 1)\mathbf{C}}{\varepsilon}}  \\
&\odot \mathbf{P}^{(\ell)}\big)  \odot \big(e ^{{\mathbf{g}^{(\ell+ 1)}}/{\varepsilon}} \odot 
\cdot \cdot  \odot e ^{{\mathbf{g}^{(1)}}/{\varepsilon}} 
\big). 
\end{split}
\label{eq:appendix_proximal}
\end{equation}
The proximal point iteration iteratively applies  Sinkhorn’s algorithm with a
$e^{-\frac{\mathbf{C}}{\varepsilon/\ell}}$
kernel, \ie \ with a decaying regularization parameter $\varepsilon/\ell$, therefore it can perform  an automatic decaying schedule on the regularization as $\ell\to\infty$ to gradually approach the optimal discrete plan.

\subsection{Sinkhorn iterations on  dual problem}
\label{sinkhorn}
\noindent
A simple approach to solving the unconstrained maximization problem in \eq{appendix_dual}  is to use an exact block coordinate ascent strategy, namely to update alternatively $\mathbf{f} $ and $  \mathbf{g}$ to cancel the respective gradients in these variables of the objective of (\ref{eq:appendix_dual}). Indeed, one can notice after a few elementary computations that, writing $\mathcal{L}_\text{dual}(\mathbf{f}, \mathbf{g}, \mathbf{s}) $ for the objective of (\ref{eq:appendix_dual}), we have the gradients, w.r.t., $\mathbf{f} $ and $  \mathbf{g}$ as follows:
\begin{align}
&\nabla_\mathbf{f}\mathcal{L}_\text{dual}(\mathbf{f}, \mathbf{g}, \mathbf{s}) =  \mathbf{a}  - 
  e^{\mathbf{f}/\varepsilon}  \odot (\mathbf{K}e^{\mathbf{g}/\varepsilon} ), \\
  &
  \nabla_\mathbf{g}\mathcal{L}_\text{dual}(\mathbf{f}, \mathbf{g}, \mathbf{s}) =  \mathbf{b}  - 
  e^{\mathbf{g}/\varepsilon}  \odot (\mathbf{K}e^{\mathbf{f}/\varepsilon} ), \label{eq:appendix_gvaraible} 
	\end{align}
 where $\mathbf{K} = e^{-{\mathbf{C}(\mathbf{s})}/{\varepsilon}}$ is defined as the Gibbs kernel in  Sinkhorn's algorithm.
 Block coordinate ascent can therefore be implemented in a closed form by applying
successively the following updates, starting from any arbitrary $\mathbf{g}^{(1)}$, for $t \geq 1$:
\begin{align}
  &\mathbf{f}^{(t+1)} = \varepsilon \ \text{log} \ \mathbf{a}  - \varepsilon \ \text{log} \ (\mathbf{K}e^{\mathbf{g}^{(t)}/\varepsilon} );  \label{eq:appen_gvaraibleupp} \\
  &\mathbf{g}^{(t+1)} = \varepsilon \ \text{log} \ \mathbf{b}  - \varepsilon \ \text{log} \ (\mathbf{K}^{\mathsf{T}}e^{\mathbf{f}^{(t + 1)}/\varepsilon} ). \label{eq:appen_gvaraible1} 
	\end{align}

\subsection{Experimental setting details}
\label{Training_details} 

\noindent \textbf{How is the $\varepsilon$  selected.}
The $\varepsilon$ controls a trade-off between exploration and exploitation. A large $\varepsilon$ leads to a greater exploration by a softer mask. We verify that soft masks converge to hard masks by training for few epochs and measuring the average of the squared differences between them for different  $\varepsilon$ candidates. As is common to hyperparameter  tuning, this requires some effort. Therefore for similar architectures we used the same $\varepsilon$, \eg, $\varepsilon = 1$ for lightweight ResNet-56 and MobileNetV2. 

\noindent \textbf{Training setting.}
In \tab{trainingsetting} we summarize the detailed training settings of this work.
For easier comparison, we 
use same training settings as ~\cite{wang2021neural} in finetuning. 
In our soft mask learning phase, we use a cosine learning rate schedule for updating the importance scores and network weights with SGD.

\begin{table*}[t]
	\centering
 	\caption{Training setting:  For the SGD optimizer, in the parentheses are the momentum and
weight decay. For ImageNet, batch size is 64 per GPU. We learn the soft mask using cosine learning rate schedule. 
    }
	\renewcommand{\arraystretch}{1.}
	\resizebox{0.99\linewidth}{!}{
	 \setlength{\tabcolsep}{15.mm}{
	   	\begin{tabular}{l|c|c|ccccc}
    	\toprule
  Datasets  & Backbone & Speedup & Pruning ratio \\
      \midrule	
		  CIFAR-10 &  
   ResNet-56 & $--$
   &[0, $p$, $p$, $p$], $p \in \{0.5, 0.7, 0.9, 0.925, 0.95\}$ 
   \\
     \midrule	
	 	 CIFAR-100  
    & VGG-19 & $--$
       & [0:0, 1-15:$p$], $p \in \{0.5, 0.6, 0.7, 0.8, 0.9\}$
    \\ 		
    \midrule	
	 	 ImageNet & ResNet-34  
    & 1.32$\times$    
    &[0, 0.50, 0.60, 0.40, 0]
    \\ 		
    \midrule	
ImageNet&  ResNet-50
    & 2.31$\times$
       &[0, 0.60, 0.60, 0.60, 0.21]
    \\ 	
    \midrule	
    ImageNet &  ResNet-50
    & 2.56$\times$
       & [0, 0.74, 0.74, 0.60, 0.21]
    \\ 	
    \midrule
    ImageNet&  ResNet-50
    & 3.06$\times$
       & [0, 0.68, 0.68, 0.68, 0.50]
    \\ 	
\bottomrule
\end{tabular}
}
}
\label{tab:Pruningratio}
\end{table*}

\noindent \textbf{Pruning ratio.}
We use same pruning ratio as in~\cite{wang2021neural} for fair comparison. 
In~\tab{Pruningratio} we give the specific pruning ratio used for our experiments in the paper.
We here briefly explain how we set the pruning ratio. We use two different architectures: single-branch (\ie VGG-19) and multi-branch (\ie ResNet). 
\textit{(i)} For VGG19, we use the following pruning ratio
setting. As an example, “[0:0, 1-9:0.3, 10-15:0.5]” means “for the first layer (index starting from 0),
the pruning ratio is 0; for layer 1 to 9, the pruning ratio is 0.3; for layer 10 to 15, the pruning ratio is
0.5”. 
\textit{(ii)} For a ResNet, if it has $N$
stages, we will use a list of $N$ floats to represent its pruning ratios for the $N$ stages. For example,
ResNet-56 has 4 stages in conv layers, then “[0, 0.7, 0.7, 0.7]” means “for the first stage (the first conv layer), the pruning ratio is set as 0; the other three stages have pruning ratio of 0.7”. Besides,
since we do not prune the last conv layer in a residual block, which means for a two-layer residual block
(for ResNet-56), we only prune the first layer; for a three-layer bottleneck block (for ResNet-34 and
ResNet-50), we only prune the first and second layers. 

\noindent \textbf{Unstructured pruning.}
For unstructured pruning, each individual weight needs to couple an importance score which may cause higher memory cost, therefore we follow previous works that use the magnitude of the weight parameter as an importance score.

\subsection{Training recipe variants}
\label{recipvariant} 
\noindent Our primary goal was to compare pruning methods as fairly as possible. The reported results for ResNet-50 used the \textit{baseline training recipe} from CReg for fair comparisons. Here, we performed additional analyses, pruning ResNet-50 on ImageNet with Label Smoothing, TrivialAugment, Random Erasing, Mixup, and Cutmix to form a \textit{stronger training recipe}. We did not apply other effective techniques such as Long Training (\eg 600 epochs), LR optimizations, EMA, Weight Decay tuning, and Inference Resize tuning, which may further improve by $\sim$3\% accuracy as shown in PyTorch. We trained on a large batch size of 4,096 on 8 NVIDIA A100 GPUs, scaling up the learning rate by 5$\times$. As expected, \tab{recipvariant} shows increased pruning accuracy, showing that our method can be augmented with different training recipes.

\begin{table}[t]
\centering
\caption{ {Training recipes
for ResNet50 under structured
pruning on ImageNet-1K. Our
method is flexible for different
training recipes.}}
\setlength{\tabcolsep}{5.mm}{
	\resizebox{0.7\linewidth}{!}{
		\begin{tabular}[c]{lc|ccccccc}
			\toprule
			{Recipe} &  {Speed
Up} &{Acc.} \\
			\midrule

   			Baseline  &\multirow{2}{*}{2.31$\times$}& 75.54  \\
			Stonger &&  77.16     \\
			\bottomrule
		\end{tabular}
	}
}	
	\label{tab:recipvariant}	
\end{table}

\begin{table}[t]
\centering
\caption{ {FLOPs budgets for WideResNet-26 on CIFAR-100}}
\setlength{\tabcolsep}{7mm}{
	\renewcommand{\arraystretch}{1}
	\resizebox{0.8\linewidth}{!}{
		\begin{tabular}[c]{lc|cccccccc}
			\toprule
			\multirow{1}{*}{Method} &  \multirow{1}{*}{Budget (\%)} & \multicolumn{1}{c}{Acc.} \\
			\midrule
   	Unpruned  & 100 &  80.21 
   \\
   \midrule			
   			ChipNet~\cite{tiwari2021chipnet}  &\multirow{2}{*}{40}&  76.88      \\
			Ours &  & 78.63       \\
   \midrule
   			ChipNet~\cite{tiwari2021chipnet}  &\multirow{2}{*}{20}&  77.15          \\
			Ours & & 78.17       \\

			\bottomrule
		\end{tabular}
	}
}	
	\label{tab:FlopsBuget}	
\end{table}

 \begin{figure}
    \centering
\includegraphics[width=0.98\linewidth]{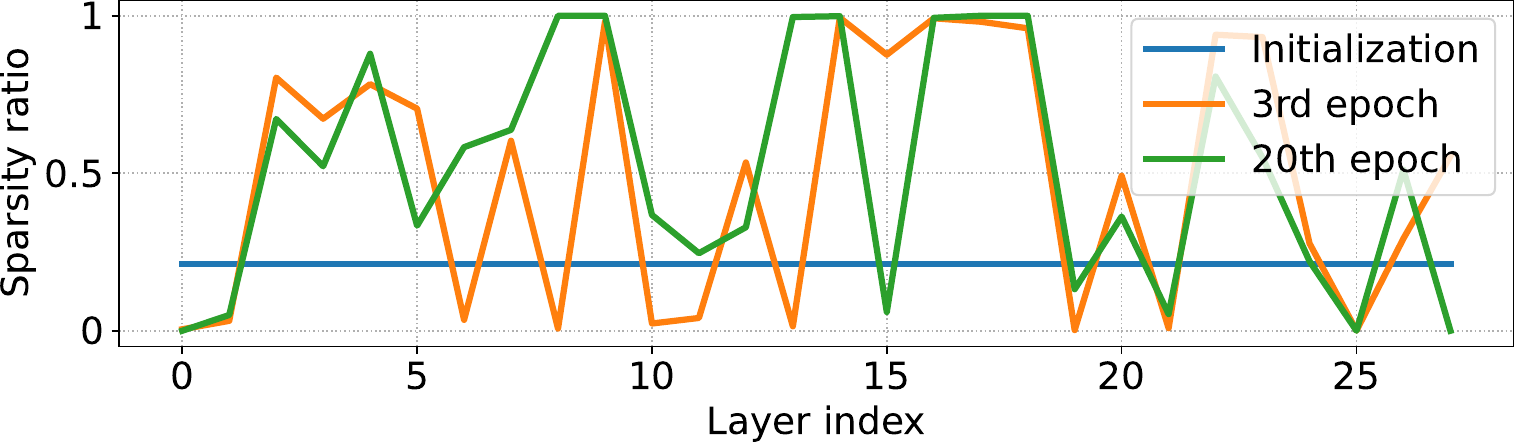}
    \caption{  Learned sparsity ratios over layers under FLOPs budget.
    }
    \label{fig:Sparsity_ratio_FLOPsbudget}
\end{figure} 

\RVC{
\subsection{Extension to FLOPs and latency budgets}
\label{flopslatency} 

\noindent
The FLOPs or latency budget pruning can be viewed as a  learnable differentiable sparsity allocation problem.
 We  formulate it in the following.  
 
\noindent \textbf{Formulation.} Given budget $B_\mathcal{F}$, 
the network weights as $\mathbf{w}$, 
the optimization problem of kept ratios 
$\mathcal{A} = \{\alpha^{(l)}\}_{l=1,\dots, L}$ 
(\ie, 1$-$sparsity ratio) of $L$
layers can be written as:
\begin{equation}
\begin{split}
& \mathop{\argmin}_{\mathbf{m}, \mathbf{w}, \mathcal{A}}\ \ 	\mathcal{L}_\text{train} \big(f(\mathbf{x};  \mathbf{w}, \mathbf{m}, \mathcal{A})\big),
 \\
 & \ \ \ \text{s.t.} \ \ \   
 \frac{1}{n} \sum \nolimits_{i=1}^n \mathbf{m}_i^{(l)} = \alpha^{(l)},
\  \   \mathbf{m}^{(l)} \in \{0, 1\}^n, 
\\
& \ \  \ \  \ \ \ \  \ \ \ \mathcal{F}(\mathcal{A}) \leq B_\mathcal{F}, \  \  0 \leq \mathcal{A} \leq 1. 
\end{split}
\label{eq:formulate_pruningnew_app}
\end{equation}
where $\mathcal{L}_\text{train}$ is   training loss. $\mathcal{F}(\mathcal{A})$ is
the consumed resource corresponding to the kept ratios. 
If $\mathcal{A}$ is predefined  layer-wise or global kept ratio, the formulation degrades to our original formulation in \eq{formulate_pruningnew}.
For FLOPs or latency budgets, we  learn the kept ratios. Note  $\mathcal{F}(\mathcal{A})$ can be denoted  as a  function of kept ratio, see~\cite{tiwari2021chipnet} and~\cite{shen2022structural}. 
By setting $\mathcal{A} = \text{Sigmoid}{(\Theta)}$ where $\Theta$ is a group of learnable parameters where $0 \leq \mathcal{A} \leq 1$ is satisfied naturally, we optimize $\Theta$ to learn kept ratio $\mathcal{A}$ by minimizng  $ \mathcal{L}_\text{train}$ and 
a budgets penalty loss $||\mathcal{F}(\mathcal{A}) - B_\mathcal{F}||^2$. Our optimal transport method dynamically aligns to  learned ratios $\mathcal{A}$.

 We add the experimental comparisons in \tab{FlopsBuget} with FLOPs budget for WideResNet-26 for CIFAR-100. 

\noindent \textbf{FLOPs budget.} 
We use the same way as~\cite{tiwari2021chipnet} to calculate the FLOPs budget that 
assumes a sliding window is used to achieve convolution and the nonlinear computational overhead is ignored. We define FLOPs budget as:
   \begin{equation}
	\mathcal{F}(\mathcal{A}) = \frac{\sum_{j = 1}^{\mathcal{N}}(K_j  \cdot n_{j-1} \cdot \alpha^{(j-1)} + 1) \cdot n_{j} \cdot  \alpha^{(j)} \cdot A_j }
 {\sum_{j = 1}^{\mathcal{N}}(K_j  \cdot n_{j-1} + 1) \cdot n_{j} \cdot A_j },
\label{eq:flops_comput}
\end{equation}
where $\mathcal{N}$ denotes the number of  convolutional layers in the network, $K_j$ denotes area of the kernel, and $A_j$ and $n_j$ denote area
of the feature maps and the channel count, respectively, in the $j^{\text{th}}$ layer.

We note that the FLOPs budget is formulated as a function of kept ratios $\alpha^{(j)}$. Also the  the latency cost of each layer  can be approximated by the kept ratios of previous layer and current layer as shown in~\cite{Humble2022pruning}.
}

\end{document}